%% file: 0main.tex
\documentclass{article}

\PassOptionsToPackage{numbers,sort}{natbib}

\usepackage[english]{babel}
    \usepackage[preprint]{neurips_2021_ml4ad}

\usepackage[utf8]{inputenc} 
\usepackage[T1]{fontenc}    
\usepackage{hyperref}       
\usepackage{url}            
\usepackage{booktabs}       
\usepackage{amsfonts}       
\usepackage{nicefrac}       
\usepackage{microtype}      
\usepackage{xcolor}         
\usepackage[pdftex]{graphicx}
\usepackage{multirow}
\newcommand{\tabincell}[2]{\begin{tabular}{@{}#1@{}}#2\end{tabular}}
\usepackage{caption}
\usepackage{algorithm}
\usepackage{algpseudocode}
\usepackage{amsmath}
\title{Hierarchical Adaptable and Transferable Networks (HATN) for Driving Behavior Prediction}

\author{%
  Letian Wang \\
  Robotics Institute\\
  Carnegie Mellon University\\
  \texttt{letian.wang@andrew.com} \\
    \And
    Yeping Hu \\
    Department of Mechanical Engineering\\
    University of California, Berkeley\\
    \texttt{yeping\_hu@berkeley.edu} \\
    \AND
    Liting Sun \\
    Department of Mechanical Engineering\\
    University of California, Berkeley\\
    \texttt{litingsun@berkeley.edu} \\
    \And
    Wei Zhan \\
    Department of Mechanical Engineering\\
    University of California, Berkeley\\
    \texttt{wzhan@berkeley.edu} \\
    \AND
    Masayoshi Tomizuka \\
    Department of Mechanical Engineering\\
    University of California, Berkeley\\
    \texttt{tomizuka@berkeley.edu} \\
    \And
    Changliu Liu \\
    Robotics Institute\\
    Carnegie Mellon University\\
    \texttt{cliu6@andrew.cmu.edu} \\}

\begin{document}

\maketitle

\begin{abstract}
  		When autonomous vehicles still struggle to solve challenging situations during on-road driving, humans have long mastered the essence of driving with efficient transferable and adaptable driving capability. By mimicking humans' cognition model and semantic understanding during driving, we present HATN, a hierarchical framework to generate high-quality driving behaviors in multi-agent dense-traffic environments. Our method hierarchically consists of a high-level intention identification and low-level action generation policy. With the semantic sub-task definition and generic state representation, the hierarchical framework is transferable across different driving scenarios. Besides, our model is also able to capture variations of driving behaviors among individuals and scenarios by an online adaptation module. We demonstrate our algorithms in the task of trajectory prediction for real traffic data at intersections and roundabouts, where we conducted extensive studies of the proposed method and demonstrated how our method outperformed other methods in terms of prediction accuracy and transferability.
\end{abstract}

\input{1introduction}

\input{3proposed_method}

\input{7experiment}

\input{8conclusion}

\input{9acknowledge}

\medskip

\bibliography{10ref}
\appendix

\input{4high_level_policy}
\input{5low_level_policy}

\input{6adaptation}
\input{7experiment_detail}
\input{11appendix}

\end{document}

%% file: 1introduction.tex
\section{Introduction}
When autonomous vehicles are deployed on the roads, they will encounter diverse scenarios varying in traffic density, road geometries, traffic rules, etc. Each scenario comes with different levels of driving difficulty. Even in a calm straight street, the sensor system of AVs still confronts daunting amount of information that may or may not be relevant to the decision making tasks. Let alone in some crazy scenarios like crowded, human-vehicle-mixed, chaotic-road-geometry intersection or roundabouts, current deployed AVs tend to timidly take conservative behaviors due to safety concern springing from insufficient capability. Nevertheless, humans can drive through and across cities so easily, even while talking to friends or shaking along the music. 

Moreover, most state-of-art decision making algorithms for AVs, once trained for one scenario, are brittle due to overspecialization and tend to fail when transferred to similar or new scenarios. On the contrary, when a fresh human driver learns how to drive in one intersection, such experience is omni-instructional and can even polish his or her driving skills in other similar or even distinct intersections and roundabouts. 

There is an obvious gap of capability between AVs and human drivers. We naturally wonder what is the secret in humans' brain, which allows us to make decisions with such ease and efficiency. Inspired by neuroscience research, our observation is that human's cheerful shuttling in dense traffic flows and complex environments benefits from two cognition mechanisms: 1) hierarchy\citep{botvinick2009hierarchically}\cite{BerliacHierachialRL2019} - cracking the entangling task into delightful sub-tasks; 2) selective attention\cite{niv2019learning}\cite{radulescu2019holistic} - identifying efficient and low-dimensional state representations among the huge information pool. Certainly, the two mechanisms are not exclusively taking effect but are highly co-related. When dividing a complex task into easier sub-tasks, humans will simultaneously choose a compact set of low-dimension states relevant to each sub-tasks. With specialized sub-policies responsible for each sub-task, humans are not only able to accomplish the task efficiently and rapidly, but also capable of generalizing their learning across scenarios due to the generic representation and reusable sub-policies.

The fruitful advantages are certainly exclusive for hierarchical methods, which end-to-end approaches\cite{salzmann2020trajectron++}\cite{codevilla2018end} cannot enjoy. However, how much we can benefit from these two mechanisms significantly depends on how properly the hierarchies and relevant states are designed. To this point, there are some existing works\cite{hu2018framework}\cite{zhao2020tnt}\cite{gao2020vectornet}
dividing the driving task into a high-level intention-determination task and a low-level action-execution task. An intention is usually defined as a goal point in the state space \cite{hu2019multi}\cite{ding2019predicting}\cite{cheng2020towards}\cite{sun2018probabilistic} or the latent space \cite{rhinehart2019precog}\cite{tang2019multiple}\cite{ding2019predicting}. Actions are then guided to reach that goal.
\begin{figure}[t]
    \centering
    \includegraphics[width=0.5\textwidth]{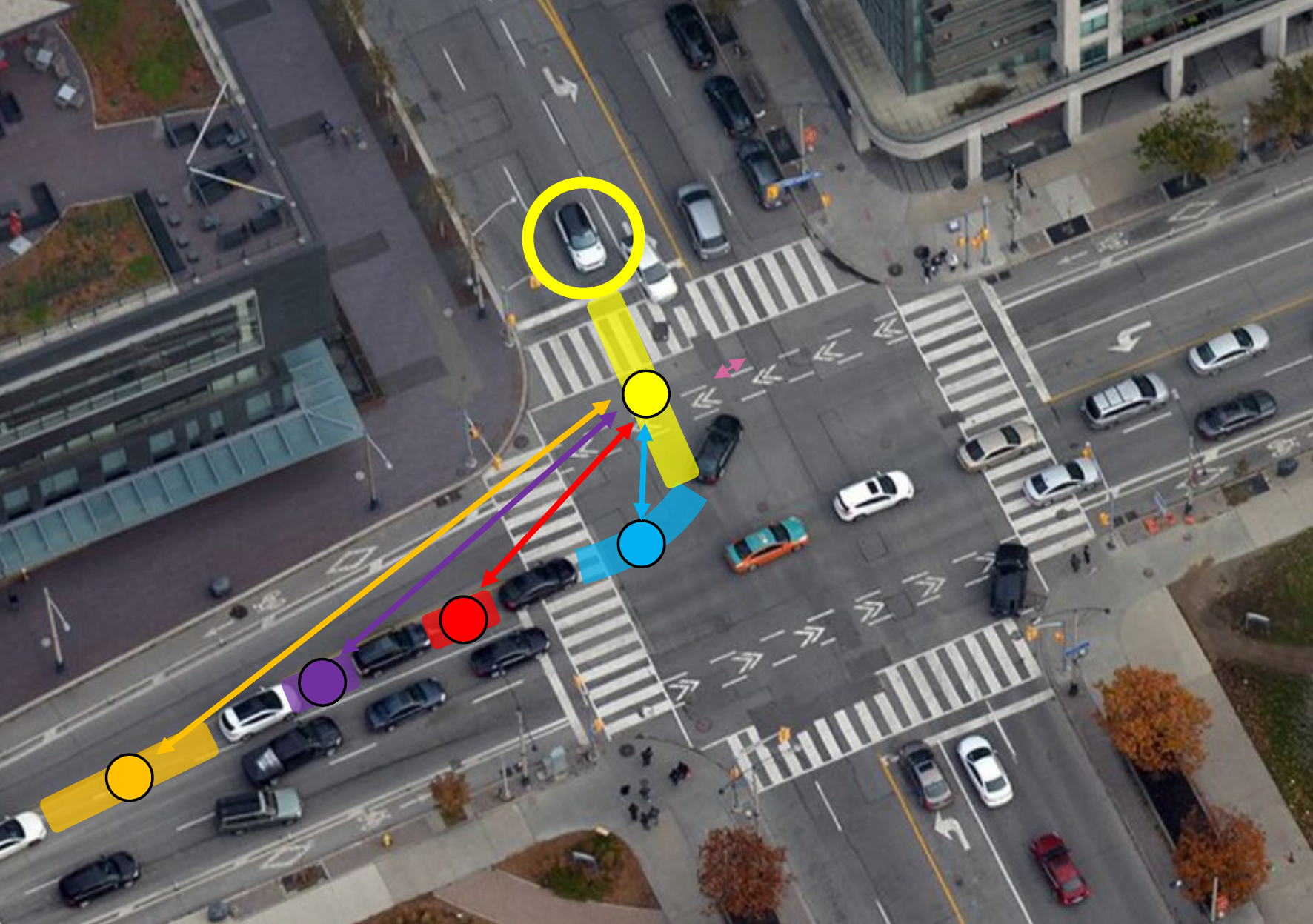}
    \caption{In dense-traffic multi-agent driving scenarios, human drivers have a hierarchical decision process: firstly decide which areas to insert into and then execute actions accordingly. Such capability is intrinsically scenario-transferable and individual-heterogeneous. Mimicking humans' driving behavior can inspire us of more advanced prediction methods.}
    \label{fig:intro}
\end{figure}

However, to gain human-level high-quality and transferable decision-making capability, the definition of hierarchy should carry more semantics by referring to how humans think while driving. On the high-level intention hierarchy, when humans are shuttling through traffic flows, they always intentionally identify which slot is most spatially and temporally proper to insert into as in Figure \ref{fig:intro}. On the low-level action hierarchy, with the chosen slot to insert into and the map geometry, humans will generate a desired reference line accordingly. Then humans will polish their micro-action skills by optimizing how well they can track the reference line. 

Such a hierarchical policy with more profound semantics enjoys many advantages. First, the policy is intrinsically scenario-transferable-and-reusable, because representation of insertion slot and reference lines can be defined consistently across scenarios. Second, the hierarchical design encourages efficient learning by reducing the size of state space for each sub-task and by polishing each sub-policy's learning individually.
 
In addition, human behaviors are naturally stochastic, heterogeneous and time-varying. For instance, humans with different driving styles\cite{wang2021socially}\cite{sun2018courteous}\cite{schwarting2019social} may result in distinct observed behaviors. Besides, though transferable, humans' behaviors are still somehow task-specific because there exist inevitable distribution shifts across scenarios, making the generalization harder. For instance, speed limit are set differently across different scenarios or cities, calling for driving customization on each scenario. Capturing such behavior variance can not only help to generate more human-like behaviors, but also encourages better generalization across scenarios. As a result, an advanced decision-making algorithm should also harness the power of online adaptation, to embrace the uncertainty in human behavior.

In summary, to generate high-quality and scenario-transferable driving behavior in multi-agent systems, we should not only design policies by leveraging human's intrinsic hierarchy and selective attention cognition model, but also capture humans' behavior variance with online adaptation methods. However, such design is not trivial. Harmonious and natural division of hierarchies, along with compact and generic state representations, are crucial to achieve what we desired. Also, to seamlessly incorporate adaptation methods into the hierarchy policies, strict mathematical formulations and systematic analysis are required.

In this paper, we propose a hierarchical, adaptable and transferable network (HATN) for high-quality and transferable behavior generation in multi-agent traffic-dense driving environments. The driving policy consists of three parts: 1) a high-level semantic graph network (SGN) responsible for slot-insertion task in multi-agent environment; 2) a low-level encoder decoder network (EDN) which executes actions according to historic dynamics and intention signal. 3) a modified extended kalman filter (MEKF) algorithm which executes online adaptation for better individual customization and scenario transfer. To the best of our knowledge, we are very first to explicitly and simultaneously take driver's nature of hierarchy, transferability and adaptability into account. 

Such a behavior generation method can be practically applied in several ways: 1) conducting accurate real-time trajectory predictions; 2) generating more human-like and user-friendly driving behaviors; 3) providing driving suggestions as an automatic driving assistance system, such as which slot to insert into; 4) reporting emergency alert when unsafe driving behavior happens. In the paper, we evaluate our method in the task of trajectory prediction due to data accessibility.

The key contributions of this paper are as follows:

\begin{enumerate}
	\item Proposed a hierarchical policy that takes in compact and generic representation, and efficiently generates human-like driving behavior in complex multi-agent intense-interaction environments, which is transferable across scenarios.
	\item Leveraged online adaptation algorithms to capture behavior variance among individuals and scenarios.
	\item Conducted extensive experiments on real human data from interaction dataset\cite{interactiondataset}, which include ablation studies for each sub-policy and show how our method outperforms other state-of-art methods in behavior forecasting task, in terms of prediction accuracy and transferability.
\end{enumerate}

%% file: 3proposed_method.tex
\section{Proposed method}
\begin{figure*}[t]
    \centering
    \includegraphics[width=0.9\textwidth]{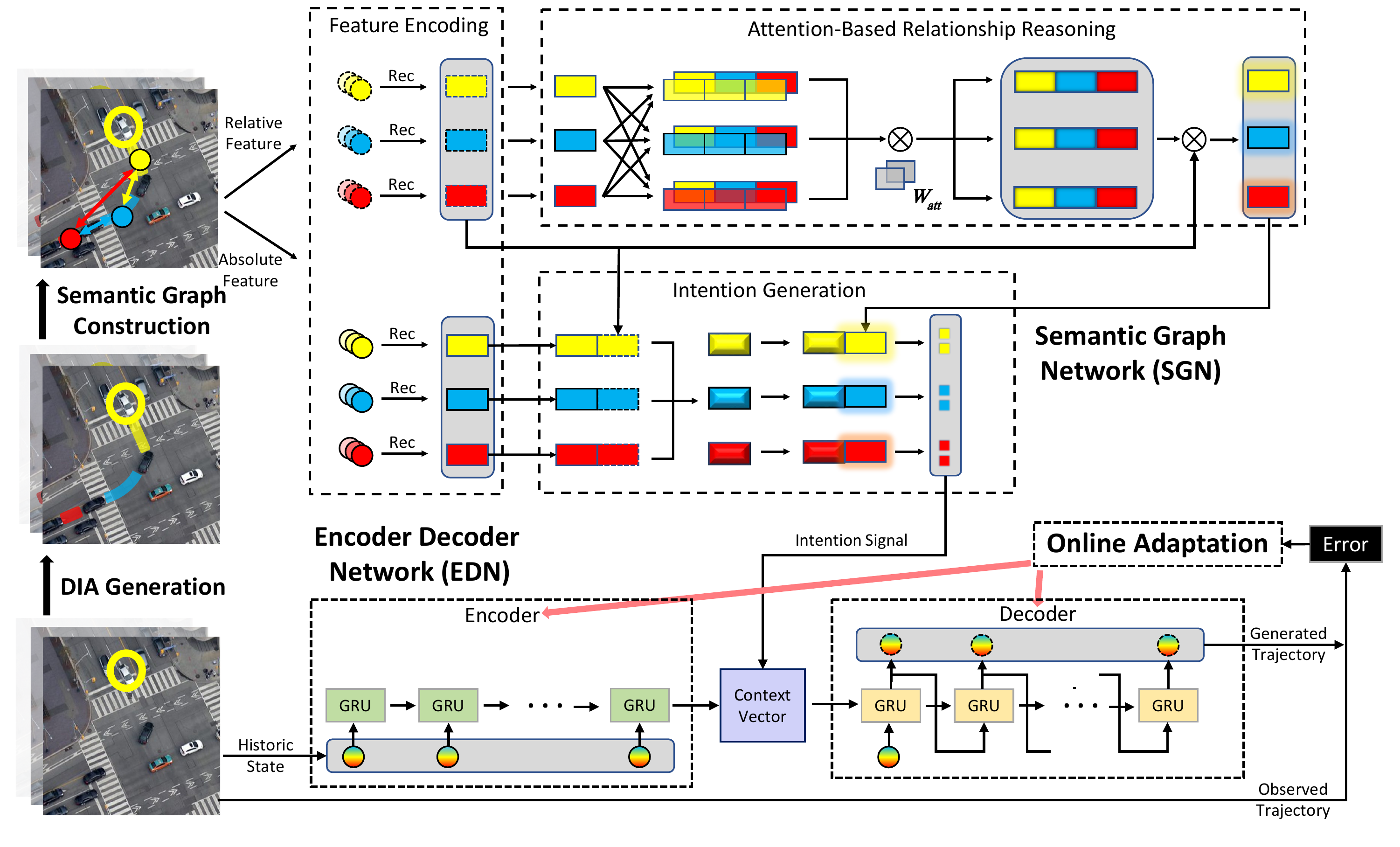}
    \caption{The proposed method consists of four parts: 1) on the left of the image, we extract ego vehicles' interacting cars and construct a Semantic Graph (SG). In SG, Dynamic Insertion Areas (DIA)\cite{hu2020scenario} are defined as the nodes of the graph, which the ego vehicle can choose to insert into. 2) Taking SG as input, the high-level Semantic Graph Network (SGN) is responsible for reasoning relationships among vehicles and predicting intention signals, such as which area to insert into and corresponding goal state. 3) The low-level Encoder Decoder Network (EDN) takes in each vehicle's historic dynamics and intention signal, and predicts their future trajectories. 4) The Online Adaptation module can online adapt EDN's parameter based on historic prediction errors, which captures individual-and-scenario-specific behaviors.}
    \label{fig:architecture}
\end{figure*}

We aim at generating more human-like driving behaviors in multi-agent traffic-dense scenarios. Specifically, with HATN, we focus on generating behaviors for any selected car and its interacting cars in the next $T_f$ seconds $\hat{\textbf{Y}}_{t + 1, t + T_f}$, based on observations of the last $T_h$ seconds $O_{t - T_h, t}$:
\begin{equation}
    \hat{\textbf{Y}}_{t + 1, t + T_f} = f_{HATN}(\textbf{O}_{t - T_h, t}).
\end{equation}

We assume perfect perception and localization conditions without vehicle-to-vehicle communications. Such assumption provides full access to the state of vehicles and is close to real driving interactions on the roads\footnote{Specifically, this paper considers intersection and roundabout scenarios, which is relatively interaction-intense, while our method can be be easily applied to other scenarios like highway and parking lot}.

Figure \ref{fig:architecture} shows the pipeline of our proposed methods. Our method essentially consists of a \textit{hierarchical policy} and an \textit{online adaptation algorithm}. The hierarchical policy is further composed of two sub-policies: a \textit{high-level intention identification policy}, and an \textit{low-level action generation policy}. Compared to monolithic policies, such a hierarchical policy can benefit from task simplification and information filtering for each hierarchy.

As mentioned previously, our insight is that when humans drive, they conceptually track a reference trajectory in their brain. The low-level action-generation policy is designed to imitate such behaviors. Intuitively, the trajectory generation procedure should certainly subject to the instantaneous dynamics of the vehicles, which requires the encoding of the historic state $\textbf{S}_{t - T_f, t}$. Furthermore, the policy should also be able to express various motion patterns, i.e. constant velocity and varying acceleration. To ensure dynamics continuity and motion diversity, we use the encoder decoder network (EDN) as the behavior-generation model, where the historic dynamics are read by the encoder part and diverse maneuvers are generated by the decoder part. With model parameter $\theta$, the task of EDN is illustrated as in the lower part of Figure~\ref{fig:architecture} and can be defined as:
\begin{equation}
    \hat{\textbf{Y}}_{t + 1, t + T_f} = f_{EDN}(\textbf{S}_{t - T_h, t}, \theta).
\end{equation}


The low-level action-generation quickly becomes ambiguous as time horizon extends, where the intention and inter-vehicle interaction come as huge impacts on driving decision. Our observation is that, when humans drive in dense traffic flows, they intuitively search for the proper slot to insert into in the intention level. Thus in the high-level intention-identification hierarchy, we first adapt a generic representation about the environment called the semantic graph (SG). In SG, we explicitly utilize dynamic insertion areas (DIA) as the node of the graph, among which the vehicles can decide to insert into or not. Such a representation is compact, efficient and generic, which captures sufficient information for intention determination and can be generically used across different driving scenarios. For more information related to DIA and SG, please refer to \cite{hu2020scenario}. Illustrated in the left part of Figure~\ref{fig:architecture}, the process of extracting semantic graph representation $\mathcal{G}_{t - T_f, t}$ from raw observations $O_{t - T_f, t}$ can also be formally described:
\begin{equation}
    \mathcal{G}_{t - T_f, t} = f_{SG}(\textbf{O}_{t - T_f, t}).
\end{equation}

With the extracted semantic graph, we then proposed a semantic graph network (SGN), which takes the semantic graph as input, inferences relationships and interactions among vehicles, and outputs the probability to insert into each dynamic insert area and the associated goal state $g_t$:
\begin{equation}
    g_t = f_{SGN}(\mathcal{G}_{t - T_f, t}).
\end{equation}

The associated goal state can be abstracted as an intention signal, which can be delivered to the low-level EDN to guide action generation:
\begin{equation}
    \hat{\textbf{Y}}_{t + 1, t + T_f} = f_{EDN}(\textbf{S}_{t - T_h, t}, g_t, \theta).
\end{equation}


Up to this point, our model is capable of generating high-level intentions and low-level actions efficiently and transferably. However, the trained model can only capture the motion pattern in a crowd sense, while the nuances among individuals can hardly be reflected. Besides, different scenarios also more-or-less customize the behavior distributions. To break the capability limit set by such behavior variance, we set up an on-line adaptation module, where a modified kalman filter (MEFK$_{\lambda}$) algorithm is used to subtly adjust model parameters for each agent based on its historic behaviors. Specifically, we regard EDN as a dynamic system and estimate its parameter $\theta$ by minimizing the error between ground-truth trajectory in the past $\tau$ steps $Y_{t-\tau, t}$ and generated trajectory $\tau$ steps earlier $\hat{\textbf{Y}}_{t-\tau,t}$: 
\begin{equation}
    \theta_{t} = f_{MEKF_{\lambda}}(\theta_{t-1}, \textbf{Y}_{t-\tau, t}, \hat{\textbf{Y}}_{t-\tau,t}).
\end{equation}

For detailed introduction of our method, please refer to Appendix~\ref{appendix:SGN}, \ref{appendix:EDN}, and \ref{appendix:adaptation}.

%% file: 7experiment.tex
\section{Experiment}
In this section, we briefly introduce the evaluation of the proposed algorithm with a case study, overall performance evaluation, and a comparison with other methods. For more detailed experiment, please refer to the Appendix \ref{appendix:experiment}. 

\subsection{Experiment setting}
We verified our proposed method with real human driving data from the INTERACTION dataset\cite{interactiondataset}. Two different scenarios were utilized as in Figure~\ref{fig:sequence of insertion} and Figure~\ref{fig:transferability}: a 5-way unsignalized intersection and an 8-way roundabout. All vehicle data were collected by a drone from bird-eye view with 10 Hz sampling frequency. Road reference paths and traffic regulations were extracted from the provided high-definition map. 

The intersection scenario was used to train our policy and evaluate the performance of behavior prediction. The roundabout scenario was used to evaluate the transferability of our method. In the intersection scenario, we had 19084 data points, which were split to 80\% of training data and 20 \% of testing data. In the roundabout scenario, there were 9711 data points to evaluate the transferability.

In our experiments, we choose the historic time steps $T_h$ as 10 and future time step $T_f$ as 30, which means we utilized historic information in the past 1 seconds to generate future behavior in the next 3 seconds. In addition to the long-term behavior prediction evaluation in the whole future 30 steps, we also evaluated short-term behavior prediction in the future 3 steps in Sec \ref{sec:overall_evaluation}, as short-term behavior is safety-critical especially in close-distance interactions.

The method was implemented in Pytorch on a desktop computer with an Intel Core i7 9th Gen CPU and a NVIDIA RTX 2060 GPU. For each model, we performed optimizations with Adam and sweep over more then 20 combinations of hyperparameter to select the best one including batch size, hidden dimension, learning rate, dropout rate, etc. 

\subsection{Case study}
\begin{figure*}[t]
    \centering
    \includegraphics[width=0.95\textwidth]{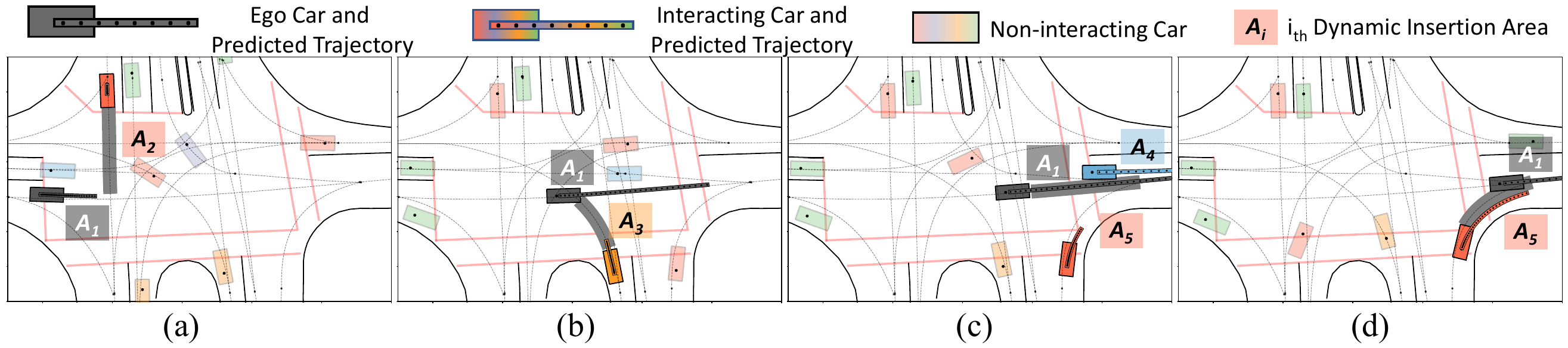}
    \caption{Case 2 - an illustration of how our method works during a sequence of interactions. When ego car crosses the interaction, our method can constantly identify interactions and extract interacting cars. The ego car first interacted with red car in (a), then with orange car in (b), then the blue and red car in (c)(d)}
    \label{fig:sequence of insertion}
\end{figure*}

We first illustrate how our method works with three examples in Figure \ref{fig: one insertion}-\ref{fig:transferability}: 1) how ego vehicle interacted with other vehicles to pass a common conflict point (one interaction); 2) how ego vehicle interacted with other vehicles to pass a sequence of conflict points (a sequence of interaction); 3) how ego vehicle interacted with other vehicles when it is zero-tranferred to the roundabout scenario (scenario-transferable interactions).
\subsubsection{Case 1: one interaction} 
\begin{figure*}[t]
    \centering
    \includegraphics[width=0.95\textwidth]{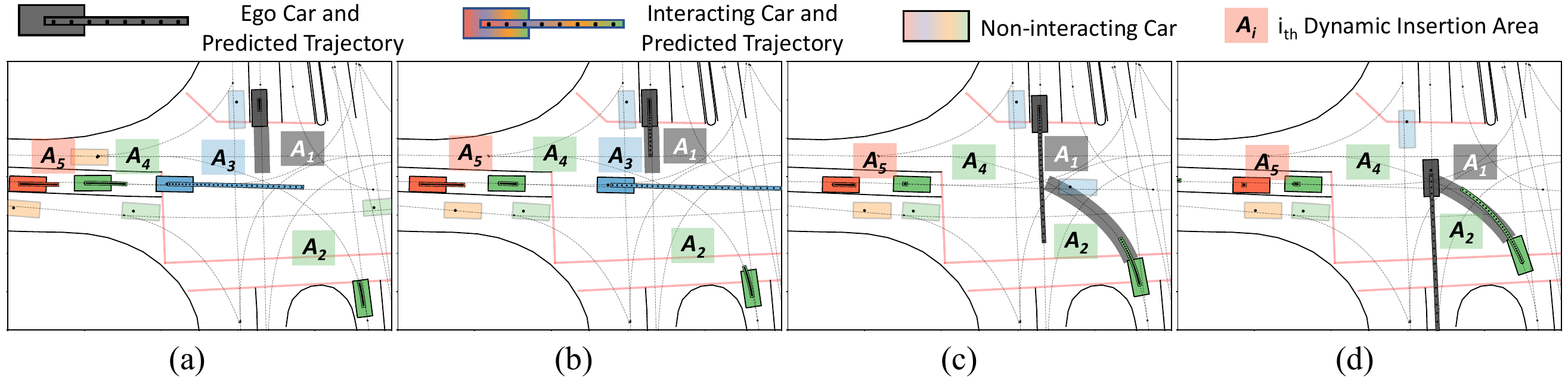}
    \caption{Case 1 - an illustration of how our method works during one interaction. Our methods can predict which DIA ego car will insert into, and the future trajectories of ego car and its interacting cars. Black car denotes the ego car; bright-color cars denote ego car's interacting vehicles, from which the DIAs are extracted. The darker one DIA is, the more likely ego car is going to insert into that area. The predicted trajectory of ego vehicle and its interacting vehicles are displayed in each vehicle's color. Transparent color cars denote the non-interacting cars. In this case, the ego car first yielded the blue car in (a)(b), and then passed before other cars (c)(d).}
    \label{fig: one insertion}
\end{figure*}

As in Figure~\ref{fig: one insertion}, we first show how our method works in one interaction. Once we chose the ego vehicle, we extracted cars whose reference lines conflict with ego car's reference line. These cars were regarded as the interacting vehicles and corresponding DIAs were extracted. Our method would then predict the intention and future trajectory of the ego vehicle and its interacting vehicles. In the figure, we drew the ego vehicle with black color, the interacting vehicles with bright colors, and the non-interacting vehicles with transparent colors. The darker one DIA is, the more likely the ego vehicle would insert into that DIA. The future trajectories of ego vehicle and interacting vehicles are also displayed with the corresponding color.

As in Figure~\ref{fig: one insertion}(a)(b), the black ego car initially had 5 areas to choose to insert into: $A_1, A_2, A_3, A_4, A_5$. At this time, the ego vehicle braked so our method predicted that it would insert into its front DIA $A_1$, which means yielding to other vehicles. In the Figure~\ref{fig: one insertion}(c), the blue vehicle behind DIA $A_3$ ran away and the ego vehicle accelerated, so our method predicted ego vehicle would insert into DIA $A_2$, which means passing before other vehicles. In Figure~\ref{fig: one insertion} (d) the ego vehicle crossed the conflict point and finished this interaction.

\subsubsection{Case 2: A sequence of interactions} We illustrate how one vehicle crossed the intersection with a sequence of interactions in Figure~\ref{fig:sequence of insertion}. Specifically, the ego vehicle initially interacted with the upper red vehicle as in Figure~\ref{fig:sequence of insertion}(a). Because the black ego vehicle was running in a high speed, our method predicted it would insert into DIA $A_2$, which means passing before the red vehicle. After finishing the first interaction, the ego vehicle then interacted with the orange vehicle below (Fig.~\ref{fig:sequence of insertion}(b)). Our method predicted the ego vehicle would continue to pass and insert into the DIA $A_3$. Later in Figure~\ref{fig:sequence of insertion}(c), the ego vehicle's path conflicted with that of the blue and red car. The ego vehicle first decelerated and our method predicted it would insert into its front DIA $A_1$ and yield other cars. In Figure~\ref{fig:sequence of insertion}(d), after the blue car ran away, the ego vehicle then accelerated and inserted into DIA $A_5$ to pass ahead of the red car.

\subsubsection{Case 3: Scenario-transferable interactions} 
\begin{figure*}[t]
    \centering
    \includegraphics[width=0.95\textwidth]{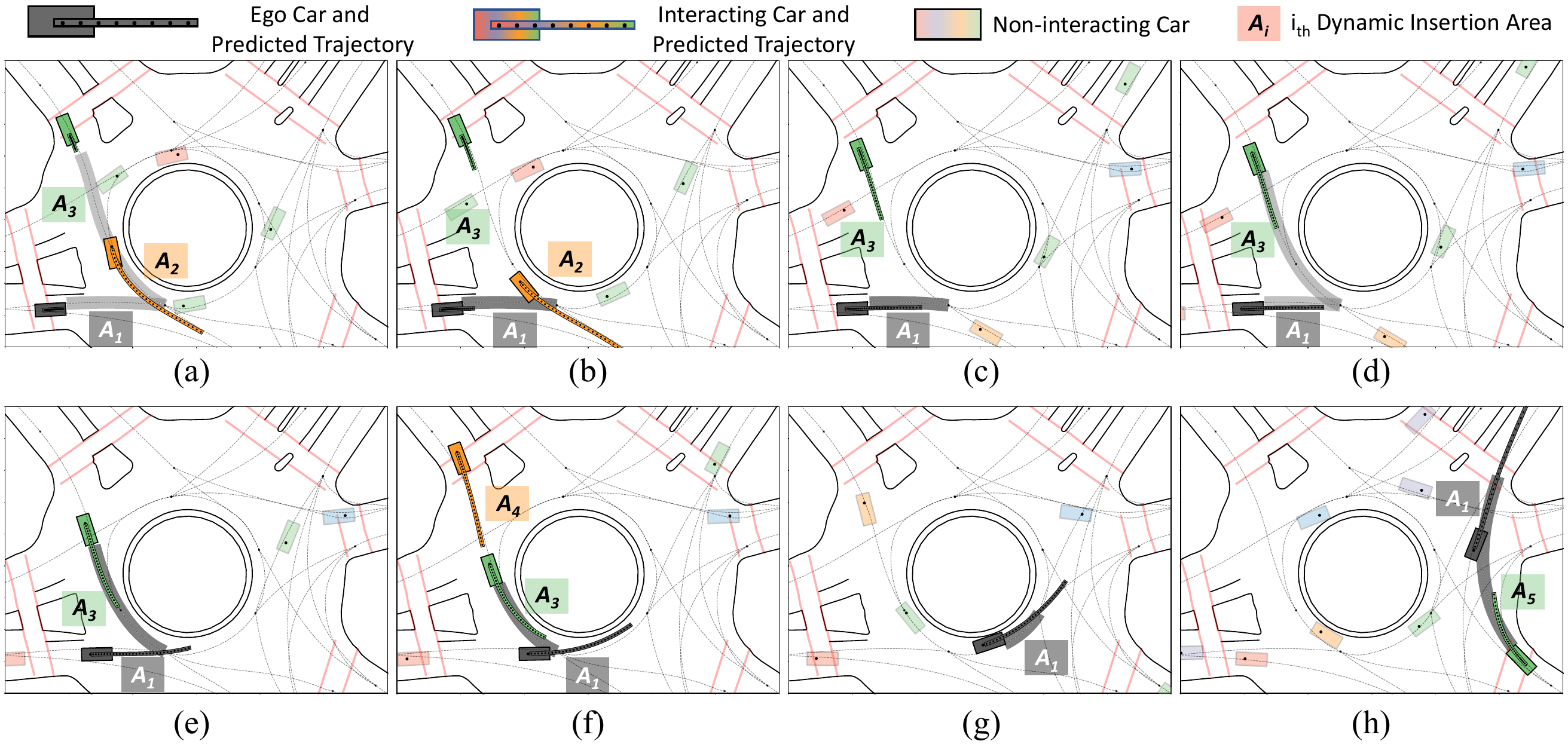}
    \caption{Case 3 - an illustration of how our method is directly transferred to the roundabout scenarios without learning a new set of parameters, after it is trained in the intersection scenario. The ego car first yielded the orange car in (a)(b), then passed before the green car (c)(d)(e)(f). Note that humans' hesitating and intention switching was captured by our method. The ego car then continued to run in (g), and left the roundabout in (h).}
    \label{fig:transferability}
\end{figure*}

In Figure~\ref{fig:transferability}, we show that after our policy was trained in the intersection scenario, it can be zero-transferred to the roundabout scenario. In Figure~\ref{fig:transferability}(a), the ego vehicle just entered the roundabout with some speed, so it was equally likely to insert into the three DIAs $A_1, A_2, A_3$. In Figure~\ref{fig:transferability}(b), the ego vehicle decelerated, which means that it would yield other vehicles and insert into its front DIA $A_1$. In Figure~\ref{fig:transferability}(c), the orange car moved away but the ego vehicle still remained a low speed, leading to the prediction that it would continue to yield. Nevertheless in Figure~\ref{fig:transferability}(d), the ego vehicle accelerated so it became equally likely to insert into either DIA $A_1$ or DIA $A_3$. In Figure~\ref{fig:transferability}(e) the ego vehicle gained a high speed, and thus our method predicted it would pass before the green car by inserting into DIA $A_3$. After finishing the first interaction in Figure~\ref{fig:transferability}(f), the ego vehicle continued to run while there were no other vehicles interacting as in Figure~\ref{fig:transferability}(g). In Figure~\ref{fig:transferability}(h), one green car entered the roundabout and the ego vehicle decided to pass before it by inserting into DIA $A_5$.

\subsection{Ablation studies}
\label{sec:overall_evaluation}
\begin{table*}[!t]
		\centering
		\caption{A brief summary of ablation studies conducted to develop our method. We incrementally investigated: the effect of the feature and representation in EDN; the benefit of intention signal to prediction; the performance of online adaptation. We evaluated the performance by calculating ADE (m) and FDE (m) between predicted trajectory and ground-truth trajectory under different horizons and scenarios.} 
		\label{tab:overall_comparison}
		\resizebox{\textwidth}{!}{%
		\begin{tabular}{ccc|c|cc|cc|c} 
			\toprule 
			& & & & \multicolumn{2}{c|}{EDN Exploration} & \multicolumn{2}{c|}{Intention Integration} & Online Adaptation\\
			\midrule 
			Scenario & Metric & Horizon & Baseline & Feature & Representation & \tabincell{c}{Goal State\\ Signal} & \tabincell{c}{Time\\Signal} & Ours\\
			\midrule 
			\multirow{4}*{Intersection} &  \multirow{2}*{ADE} & 3s  & 0.884 ± 0.594 & 0.629 ± 0.397 & 0.407 ± 0.328 & 0.302 ± 0.251 & 0.305 ± 0.253 & \textbf{0.301 ± 0.250}\\
			~ & ~ & 0.3s & 0.409 ± 0.245 &  0.319 ± 0.224 &  0.027 ± 0.020 &  \textbf{0.021 ± 0.017} &  0.029 ± 0.020 & 0.023 ± 0.014\\
			\cmidrule(r){4-9}    
			~ & \multirow{2}*{FDE} & 3s & 1.850 ± 1.684 & 1.416 ± 1.175 & 1.279 ± 1.130 & 0.890 ± 0.831 & \textbf{0.876 ± 0.835} & 0.877 ± 0.830\\
			~ & ~ & 0.3s & 0.423 ± 0.245 &  0.324 ± 0.229 &  0.040 ± 0.034 &  0.036 ± 0.025 &  0.043 ± 0.032 & \textbf{0.032 ± 0.024}\\
			\midrule 
			
			\multirow{4}*{\tabincell{c}{Roundabout\\(Transfer)}} &  \multirow{2}*{ADE} & 3s & 2.924 ± 4.695 & 2.201 ± 3.999 & 0.941 ± 0.778 & 0.845 ± 0.564 & 0.815 ± 0.526 & \textbf{0.815 ± 0.526}\\
			~ & ~ & 0.3s & 1.572 ± 4.029 & 1.543 ± 3.957 & 0.062 ± 0.071 & 0.060 ± 0.060 & 0.073 ± 0.065 & \textbf{0.052 ± 0.068} \\
			\cmidrule(r){4-9}    
			
			~ &	\multirow{2}*{FDE} & 3s & 5.446 ± 7.157 & 4.123 ± 5.227 & 2.494 ± 2.070 & 2.081 ± 1.440 & 2.038 ± 1.409 & \textbf{2.041 ± 1.409} \\
			~ & ~ & 0.3s & 1.546 ± 3.894 & 1.544 ± 3.944 & 0.088 ± 0.104 & 0.091 ± 0.101 & 0.108 ± 0.107 & \textbf{0.079 ± 0.114}\\
			\bottomrule 
		\end{tabular} %
		}
		
\end{table*}

In this section, we briefly summarize all the experiments conducted as a quick take-away for the reader. Table~\ref{tab:overall_comparison} shows the distilled results of all the ablations studies to explore the effect of different factors. 

Starting from a baseline EDN which takes in the historic position and predicts future trajectory, we first explored the effect of features and representations in EDN. By adding features such as speed and yaw, we reduced the ADE in 3 seconds by 28\% and 24\% in the two scenarios. In the representation aspect, we conducted incremental prediction and position alignment. Such design not only effectively reduced the ADE in 3 seconds by 35\% and 57\% in the two scenarios, but also significantly reduced the ADE in 0.3 seconds by 91\% and 95\%. We can consequently conclude that the information of speed and yaw, along with the incremental prediction and position alignment, are vital for the prediction performance in encoder decoder architecture. 

\begin{figure*}[t]
    \centering
    \includegraphics[width=1\textwidth]{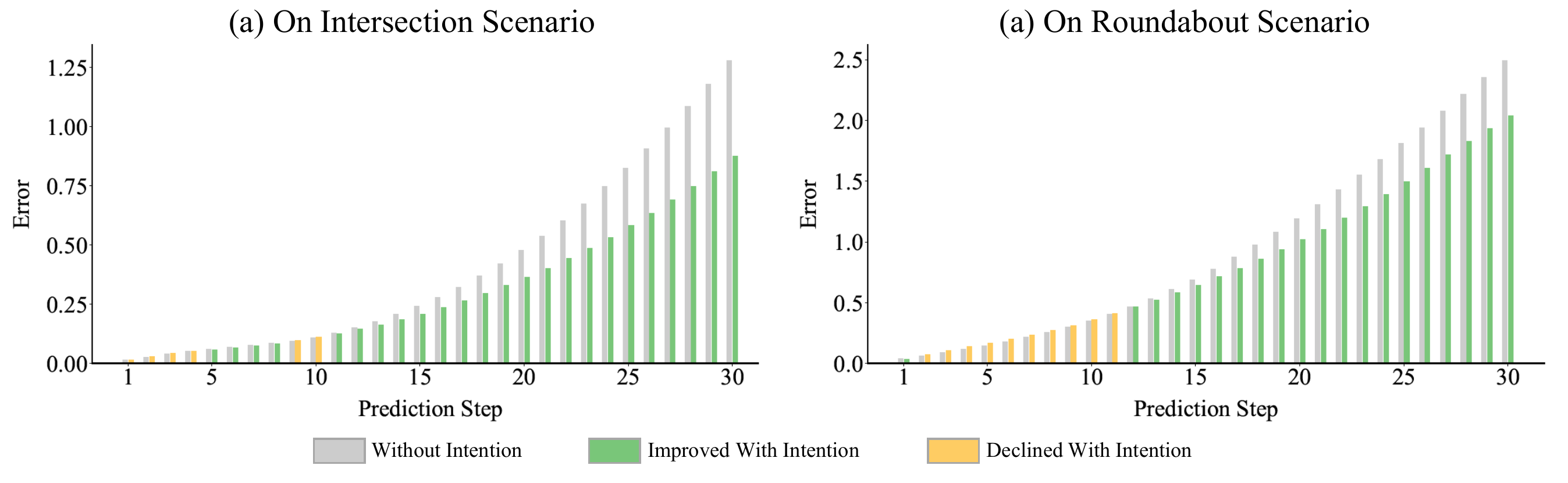}
    \caption{Prediction error (m) of each step. The prediction error grew exponentially as horizon extends. The intention signal effectively suppressed the error growth, especially in the long term.}
    \label{fig:Error by Step}
\end{figure*}

Next, we integrated intention signals into the EDN. According to the Table~\ref{tab:overall_comparison}, integrating the goal state signal significantly reduced the ADE in 3 seconds by 25\% and 10\% in the two scenarios. The time signal barely benefitted the prediction in the intersection scenario, but it reduced the ADE in 3 seconds of roundabout scenario by 4\%. In Figure~\ref{fig:Error by Step}, we also displayed the error by step in the future 30 prediction steps. We can clearly see that as the prediction horizon extended, the prediction error grew exponentially. But the intention signal effectively suppressed the error growth, especially in the long horizon. Such results effectively demonstrated the necessity of intention integration.

Finally, the online adaptation was implemented. It can be seen that the online adaptation barely improved the long-term prediction in the next 3 seconds, but the short-term prediction in 0.3 seconds was improved by 20\% and 28\% in the intersection and roundabout scenarios. As a result, a conclusion is that though the online adaptation may not help with long-term prediction, but it can help to refine short-term prediction. Such improvement is valuable especially in close-distance prediction, where small prediction shift can make a big difference in terms of safety.

\subsection{Comparison with other methods}
In this section, we compared our method with other methods in terms of behavior prediction accuracy in different horizons and scenarios.

As in Table~\ref{tab:SOTA}, we first considered three rule-based methods. The IDM\cite{treiber2000congested} method basically follows its front car on the same reference path. The FSM-based\cite{zhang2017finite} method will also recognize cars on the other reference paths, and follow the closest front car using the IDM model. When deciding the closest front car, the FSM-D method will compare ego vehicle's distance to the reference line conflict point to that of other cars, and choose the closest car. The FSM-T method will first calculate each vehicle's time needed to reach the reference line conflict point by assuming they are running in constant speed, and then choose the closest one. We set the parameter for the IDM model identified in urban driving situations\cite{liebner2012driver}.

As shown in Table~\ref{tab:SOTA}, the rule-based method had not been working well in the prediction task. Several reasons may be possible. First is that intersections and roundabouts are really complicated with intense multi-agent interactions. Simple rules can hardly capture such complex behaviors while systematic rules are hard to manually design. A second reason is that, the parameter in the driving model is hard to specify as it is also scenario-and-individual-specific.

\begin{table*}[!t]
		\centering
		\caption{Performance comparison among rule-based method, state-of-the-art method, and our method. We evaluated the performance by calculating ADE (m) and FDE (m) between predicted trajectory and ground-truth trajectory in different horizons and scenarios.} 
		\label{tab:SOTA}
		\resizebox{\textwidth}{!}{%
		\begin{tabular}{ccc|ccc|cc} 
			\toprule 
			& & & \multicolumn{3}{c|}{Rule-Based Method} & \multicolumn{2}{c}{GNN-Based Method}\\
			\midrule 
			Scenario & Metric & Horizon & IDM & FSM-D & FSM-T & Trajectron++ & Ours\\
			\midrule 
			\multirow{4}*{Intersection} &  \multirow{2}*{ADE} & 3s  & 2.847 ± 1.963 & 3.181 ± 2.403 & 3.372 ± 2.495 & 0.510 ± 0.440 & \textbf{0.301 ± 0.250}\\
			~ & ~ & 0.3s & 0.042 ± 0.014 & 0.051 ± 0.034 & 0.054 ± 0.034 & \textbf{0.009 ± 0.008} & 0.023 ± 0.014\\
			\cmidrule(r){4-8}    
			~ & \multirow{2}*{FDE} & 3s & 7.655 ± 5.536 & 8.709 ± 7.081 & 9.011 ± 7.192 & 1.617 ± 1.517 & \textbf{0.877 ± 0.830}\\
			~ & ~ & 0.3s & 0.091 ± 0.033 & 0.103 ± 0.059 & 0.111 ± 0.059 &  \textbf{0.014 ± 0.013} & 0.032 ± 0.024\\
			\midrule 
			
			\multirow{4}*{\tabincell{c}{Roundabout\\(Transfer)}} &  \multirow{2}*{ADE} & 3s & 5.271 ± 1.950 & 4.637 ± 1.448 & 4.824 ± 1.509 & 1.250 ± 0.849 & \textbf{0.815 ± 0.526}\\
			~ & ~ & 0.3s & 0.126 ± 0.062 & 0.090 ± 0.071 & 0.093 ± 0.070 & \textbf{0.015 ± 0.011} & 0.052 ± 0.068\\
			\cmidrule(r){4-8}    
			
			~ &	\multirow{2}*{FDE} & 3s & 13.891 ± 5.845 & 13.133 ± 4.208 & 13.505 ± 4.407 & 4.063 ± 2.706 & \textbf{2.041 ± 1.409}\\
			~ & ~ & 0.3s & 0.206 ± 0.096 & 0.157 ± 0.105 & 0.162 ± 0.102 & \textbf{0.025 ± 0.020} & 0.079 ± 0.114\\
			\bottomrule 
		\end{tabular} %
		}
		
\end{table*}

We also implemented trajectron++\cite{salzmann2020trajectron++}, a state-of-the-art GNN-based method. This method takes vehicles as nodes of a graph and utilized a graph neural network to conduct relationship reasoning. The map information is integrated by embedding the image of the map. A generative model is then used to predict future actions. Future trajectory is then generated by propagating the vehicle dynamics with the predicted actions.

According to Table~\ref{tab:SOTA}, our method significantly outperformed trajectron++, with ADE in 3 seconds lower by 41\% and 34\% in the intersection and roundabout scenarios respectively. Such results demonstrated that our method's great capability in the long-term prediction, benefiting from our design of semantic hierarchy and transferable representation. Nevertheless, trajectron++ performed better than ours in the short-term prediction. One important reason is that the predicted trajectory of trajectron++ is strictly dynamics-feasible, as it will essentially predict future actions and propagate them through the dynamics to retrieve future trajectories. Such results motivate us to include dynamic constrain in our future work.

%% file: 8conclusion.tex
\section{Conclusion}
In this paper, we proposed a hierarchical framework to generate high-quality driving behavior in multi-agent environment, which is scenario-transferable and individual-adaptable. Our method demonstrated state-of-the-art performance in the trajectory prediction task based on thorough evaluations under real-world driving scenarios. In the future, it is interesting to explore the interaction between high-level policy and low-level policy to accommodate for each other, simultaneous online adaptation for intention and action, and more systematic comparisons with other methods.

%% file: 9acknowledge.tex
\section*{Acknowledgment}
The authors would like to thank Abulikemu Abuduweili and Alvin Shek for insightful and helpful discussions.

%% file: 4high_level_policy.tex
\section{High-level intention-identification policy}
\label{appendix:SGN}
In this section, we first
adopt the semantic graph (SG), which is originally introduced in \cite{hu2020scenario}. In the SG, the dynamic insertion area (DIA) is
defined as a generic and compact representation of the scenario. We then introduce the semantic graph network
(SGN) which generates agents' intention by reasoning about their internal relationships. The advantages of adopting Dynamic Insertion
Area can be threefold: (1) It explicitly describes humans' insertion behavior considering map, traffic regulations, and interaction information. (2) It filters scene information and only extracts a compact set of vehicles and states crucial for the intention prediction task. (3) DIA is a generic representation, which can be used across different scenarios. 

\subsection{Semantic graph}
\label{Sec:Semantic Graph}
The semantic graph (SG) utilize Dynamic Insertion Area (DIA) as basic nodes for a generic spatial-temporal representation of the environment. Specifically, DIA is defined as: \textit{a dynamic area that can be inserted or entered by agents on the roads.} Mathematically, we define each dynamic insertion area as $\mathcal{A} = (\mathcal{X}_f, \mathcal{X}_r, \mathcal{X}_{ref})$. $\mathcal{X}_{f/r} = (x_{f/r}, y_{f/r}, v_{f/r}, \theta_{f/r} )$ denotes the properties for the center point of front bound and rear boundary of the DIA (location, velocity and orientation); $\mathcal{X}_{ref}$ denotes the information of the reference path that the area is currently on, where the property of side boundary is also explicitly incorporated. 
To capture each DIA's crucial information for humans' decision, we further extract three high-level features under the Frenet coordinate: $d_{f/r}^{lon}$ denotes longitudinal distance to active conflict point $rpt_{act}$ of front or rear boundary. $l = d_r^{lon} - d_f^{lon}$ measures the length of the DIA. To facilitate relationship inference among DIAs, we also define the relative feature for each DIA by aligning it with the reference DIA. Note that we choose front DIA as the reference DIA because the ego vehicle is implicitly represented by the rear boundary of the front DIA. 

With the extracted DIAs, the 3D spatial-temporal semantic graph ${\mathcal{G}^{t-T_h\rightarrow t}} = ({\mathcal{N}^{t-T_h\rightarrow t}}, {\mathcal{E}^{t-T_h\rightarrow t}})$ can be constructed, where $t-T_h\rightarrow t$ denotes the time span from a previous time step $t-T_h$ to current time step $t$ with $T_h$ denoting the horizon. For more detailed description on the DIA properties, DIA extraction algorithm, feature selection, semantic graph construction, please refer to \cite{hu2020scenario}\cite{hu2018probabilistic}.

\subsection{Semantic graph network}
As the high-level intention identification policy, the architecture of the SGN is shown in Figure \ref{fig:architecture}. SGN takes the spatial-temporal 3D semantic graph from historic time step $t-T_h$ to current time step $t$ as the input, rather than only spatial 2D semantic graph of current time step $t$ in \cite{hu2020scenario}. Such a change aims at capturing more temporal dynamics and interactions among vehicles. SGN then decides which area to insert into and generate the associated goal state distribution. The mean of the goal state distribution is then delivered to low-level policy for generating more human-like behaviors.

\subsubsection{Feature encoding layer}
In this layer, we essentially encode the absolute and relative features for each node from historic time step $t-T_h$ to current time step $t$:
\begin{equation}
	h^t_i = f^1_{rec}(\textbf{X}^t_i)
\end{equation}
\begin{equation}
	h{'}^t_i = f^2_{rec}(\textbf{X}{'}^t_i)
\end{equation}
where $\textbf{X}^t_i=[x^{t-T_h}_i, ..., x^{t}_i]$ and  $\textbf{X}{'}^t_i=[x{'}^{t-T_h}_i, ..., x{'}^{t}_i]$ respectively denotes the absolute feature and relative feature of node $i$ from time step $t-T_f$ to current time $t$; $h^t_i$ and $h{'}^t_i$ denote the hidden states encoded from absolute and relative features respectively, namely the outputs of the recurrent function $f^1_{rec}$ and $f^2_{rec}$. $h^t_i$ and $h{'}^t_i$ are further embedded for later use.
\begin{equation}
	\hat{h}^t_i = f^1_{enc}(h^t_i)
\end{equation}
\begin{equation}
	\hat{h}{'}^t_i = f^2_{enc}(h{'}^t_i)
\end{equation}

\subsubsection{Attention-based relationship reasoning layer}
\label{sec:Relationship Reasoning Layer}
To infer relationships between any two nodes, inspired by Graph Attention Network\cite{velivckovic2017graph}, we design an attention-based relationship reasoning layer. In this layer, we exploit the soft-attention mechanism\cite{luong2015effective}\cite{bahdanau2014neural} to compute the node $n_i$'s attention coefficients on node $n_j$:
\begin{equation}
	a^t_{ji} = f_{att}(concat(\hat{h}{'}^t_j, \hat{h}{'}^t_i); \textbf{W}_{att})
\end{equation} 
where function $f_{att}$ maps each concatenated two features into a scalar with the parameter $\textbf{W}_{att}$. The attention coefficient is then normalized across all nodes $\mathcal{N}^t$ at time step $t$:
\begin{equation}
	\alpha^t_{ji} = \frac{exp(a^t_{ji})}{\sum_{n\in \mathcal{N}^t}exp(a^t_{ni})}
\end{equation}
Eventually, node $i$'s relationships with all nodes in the graph (including node $i$ itself) is derived by the attention-weighted summation of all encoded relative features:
\begin{equation}
	\bar{h}^t_i = \sum_{n\in \mathcal{N}^t} \alpha^t_{ni} \odot \hat{h}{'}^{t}_{n}
\end{equation} 
where $\odot$ denote element-wise multiplication.

\subsubsection{Intention generation layer}
When predicting the intention, in addition to the node relationships, each node's own features are also required. Thus we first concatenate and encode each node's embedded absolute and relative feature:   
\begin{equation}
	\widetilde{h}^t_i =  f^2_{enc}(concat(\hat{h}^t_i, 	\hat{h}{'}^t_i))
\end{equation}
Each DIA's future evolution in the latent space is then derived by combining encoded node relationships and features:
\begin{equation}
	z_i^t = f^3_{enc}(concat(\bar{h}^t_i, \widetilde{h}^t_i))
\end{equation} 
In this paper, the intention is defined as which DIA the ego car decides to insert into.
Thus the latent vector representing each DIA's evolution is then used to generate the probability of being inserted by ego vehicle: 
\begin{equation}
	\label{eq:insert_probability}
	w_i^t = \frac{1}{1 + exp(f^1_{out}(z_i^t))}
\end{equation} 
which is then normalized across all DIAs in current time step such that $\sum_{i \in \mathcal{N}^t}w_i^t = 1$.
Moreover, to generate a practical intention signal for low-level policy to leverage, we further use Gaussian Mixture Model (GMM) to generate probabilistic distribution of each DIA's future goal state $g$ in a certain horizon\footnote{In this paper, we use the relative traveled distance in future 3 seconds as the goal state representation}:
\begin{equation}
	f(g_i^t|z_i^t) = f(g_i^t|f^2_{out}(z_i^t))
\end{equation}
where the function $f^2_{out}$ maps the latent state $z_i^t$ to the parameters of GMM (i.e. mixing coefficient $\alpha$, mean $\mu$, and covariance $\sigma$). The goal state $g$ then can be retrieved by sampling in the GMM distribution. 

\subsubsection{Loss function}
We not only expect the largest probability to be associated with the actual inserted area ($\mathcal{L}_{class}$), but also the ground-truth goal state to achieve the highest probability in the output distribution ($\mathcal{L}_{regress}$). Thus we define the loss function as:
\begin{equation}
\begin{split}
	\mathcal{L} &= \mathcal{L}_{regress} +\beta \mathcal{L}_{class}\\
				&= - \sum_{\mathcal{G}_s} \bigg(\sum_{i \in \mathcal{N}^s}log\Big(   p\big(\check{g}_i|f^2_{out}(z_i)\big)\Big)+\beta \sum_{i \in \mathcal{N}^s} \check{w}_i log(w_i)\bigg)
\end{split}
\end{equation}
where $\mathcal{G}_s$ denotes all the training graph samples; $\mathcal{N}^s$ denotes all the nodes in one training graph sample; $\check{g}_i$ and $\check{w}_i$ denotes the ground-truth label for goal state and insertion probability of node $n_i$. Though our goal is to predict ego vehicle's future motion, we output the goal state for all interacting vehicles rather than ego vehicle exclusively\cite{hu2020scenario} to encourage sufficient relationship and interaction reasoning, and also realize data augmentation. 

Also note that though not directly used by low-level behavior-generation policy, the insertion probability learning $\mathcal{L}_{class}$ can serve as an auxiliary task to stabilize main-stream goal state learning $\mathcal{L}_{regress}$\cite{mirowski2016learning}\cite{hasenclever2020comic}. Defining goal state in the state space instead of the latent space also offers us accessible labels to monitor the high-level policy learning. The detailed introduction of the layers can be found in Table~\ref{tab:GNN table} of the appendix.

%% file: 5low_level_policy.tex
\section{Low-level behavior-generation policy}
\label{appendix:EDN}
To generate future behaviors of arbitrary length, we use the encoder-decoder network (EDN)\cite{cho2014learning}\cite{neubig2017neural} as the low-level behavior-generation policy, given the historic information and intention signal from high-level policy.

\subsection{Encoder decoder network}
The EDN consists of two GRU networks, namely the encoder and the decoder. At any time step $t$, the encoder takes in the sequence of historic and current vehicle states  $\textbf{S}_{t} = [\textbf{s}_{t-T_h},... \textbf{s}_{t}]$, and compresses all these information into a context vector $\textbf{c}_t$. The context vector is then fed into the decoder as the initial hidden state to recursively generate future behaviors $\hat{\textbf{Y}}_t=[\hat{\textbf{y}}_{t+1}, ..., \hat{\textbf{y}}_{t+T_f}]$. Specifically, the decoder takes vehicle's current state as the initial input to generate the first-step behavior. In every following step, the decoder takes the output value of last step as the input to generate a new output. Mathematically, the relationship among encoder, decoder, and context vector can be compactly described:
\begin{equation}
\label{eq:encoder}
	\textbf{c}_t = f_{enc}(\textbf{S}_t; \theta^{E})
\end{equation}
\begin{equation}
\label{eq:decoder}
	\hat{\textbf{Y}}_t = f_{dec}(\textbf{c}_t, s_t; \theta^{D})
\end{equation}
where the context vector $\textbf{c}_t$ is indeed the last hidden state of encoder and also used as the decoder's initial hidden state; the current state $\textbf{s}_t$ is fed as the decoder's first-step input. In this paper, we choose graph recurrent unit (GRU) as the basic RNN cell and stack three dense layers on the decoder for stronger decoding capability.

The goal of EDN is to minimize the error between ground-truth trajectory and generated trajectory. Taking a deterministic approach, the loss function can be simply designed:
\begin{equation}
    \mathcal{L} = \sum_{i=0}^N||\hat{\textbf{Y}}_i - \textbf{Y}_i||_p
\end{equation}
where $N$ denotes the number of training trajectory samples. The objective can be measured in any $l_p$ norm, while in this paper we consider $l_2$ norm.

\subsection{Integrating the intention signal}
\label{sec:intention_signal_intro_method}
The EDN can be simply regarded as a motion generator given the historic dynamics, while the encoding for interaction and map information are left to the high-level intention policy to take care. Such a hierarchical policy simplifies learning burden for each sub-policy and offers better interpretability. However, it remains unknown \textit{what} intention signals should be considered and \textit{how} to integrate them into the low-level policy.

In our case, we aim at generating more human-like behaviors in a certain future horizon. So in the intention signal we naturally expect the goal state in the future horizon to guide the EDN's generation process. Besides, considering the fact that the same GRU cell is recursively utilized at each step, the GRU cell may be confused about whether the current decoding lies in the earlier horizon or the later horizon. Consequently, we introduce the current decoding step as another intention signal to enlighten the decoder to better track the goal state. Introducing the intention signal would then modifies the decoder definition as in Eq \eqref{eq:decoder}:
\begin{equation}
	\hat{\textbf{Y}}_t = f_{dec}(\textbf{c}_t, s_t, g_t; \theta^{D})
\end{equation}

There are various ways to incorporate additional features into the time series model. When the additional feature is a temporal series, it is intuitive to append it to the end of original input feature vector or output vector of the GRU (before the dense layers) \cite{cheng2020towards}\cite{cheng2019human}. However, when we have a non-temporal-series additional feature, directly appending it to the original feature vector may create harder learning by polluting the temporal structure. A more delicate approach is to embed the additional feature with a dense layer and add it to the hidden state of RNN at the first-step decoding, so that the non-temporal signal is passed in the GRU cell state along the decoding sequence\cite{karpathy2015deep}\cite{vinyals2015show}. Besides, in our case, the goal state intention signal is defined as the goal state in the physical world, so another approach is to directly transform the original input state to the state relative to goal state, such that the model is implicitly told to reach origin position at the last step of decoding. 

Additionally, the intention signal from the high-level policy may itself carry the error variance springing from the policy structure or data distribution. Thus in this paper, we will systematically experiment for different input features, data representation, and intention signal introduction methods in Sec.\ref{sec:experiment_intention_signal_intro}.

%% file: 6adaptation.tex
\section{Online adaptation}
\label{appendix:adaptation}
Humans are born to be irrational, resulting in heterogeneous, stochastic, and time-varying behaviors. Task structures also inevitably create additional behavior shift. We thus utilize online adaptation to inject customized individual and scenario patterns into the model. 

\subsection{Multi-step feedback adaptation formulation}
The goal of online adaptation is to improve the quality of behavior prediction with the feedback from the historic ground-truth information. In our case, the policy is hierarchically divided, with two sub-policies to be adapted. However, the online adaptation of high-level intention-identification policy has been hard as the ground-truth intention is not directly observable in real time. Thus in this paper, we only consider the online adaptation for the low-level behavior-generation policy, while keeping the high-level intention-identification policy intact. The intuition behind the online adaptation is thus that, though given the same goal state, drivers still have diverse ways to achieve it. Capturing such customized pattern can improve the human-likeness of generated behavior. 

Formally, at time step $t$, online adaptation aims at exploiting local over-fitting to improve individual behavior generation quality: 
\begin{equation}
    \min_{\theta} ||\hat{\textbf{Y}}_t - \textbf{Y}_t||_p
\end{equation}
where $\textbf{Y}_t$ is the ground-truth trajectory; $\hat{\textbf{Y}}_t$ is the generated future trajectory by the EDN with the model parameter $\theta$. Assume that the model parameter changes slowly, namely $\dot{\theta}\approx0$. Then the model parameter that generates the best predictions in the future can be approximated by the model parameter that fits the historic ground-truth observation best. Also note that online adaptation can be iteratively executed when new observation is received.

Practically, the length of ground-truth observation $\tau$ may not necessarily match the behavior generation horizon $T_f$, as online adaptation can be conducted as soon as at least one-step new observation is available. Thus the online adaptation is indeed a multistep feedback strategy\cite{abuduweili2021robust}. By definition, at time step $t$, we have the recent $\tau$ step ground-truth observation $\textbf{Y}_{t-\tau, t} = [y_{t-\tau+1}, y_{t-\tau+2}, ..., y_{t}]$. From the memory buffer we also have the generated behavior at $t-\tau$ steps earlier $\hat{\textbf{Y}}_{t-\tau, t} = [\hat{y}_{t-\tau+1}, \hat{y}_{t-\tau+2}, ..., \hat{y}_{t}]$. Then the model parameter is adapted based on the recent $\tau$-step error:
\begin{equation}
    \hat{\theta}_t = f_{adapt}(\hat{\theta}_{t-1}, \hat{\textbf{Y}}_{t-\tau, t}, \textbf{Y}_{t-\tau, t})
\end{equation}

The adapted model is then used to generate behaviors in the future $T_f$ steps from current time $t$. $f_{adapt}$ denotes the adaptation algorithm to be discussed in detail in Sec.\ref{sec:MEKF}. It worths noting that here only $\tau$-steps errors are utilized and we expect better performance in $T_f$ steps. Nevertheless, behavior prediction accuracy usually decays exponentially as horizon extends. When $\tau$ is small, we may not obtain enough information for the online adaptation to benefit behavior prediction in the whole future $T_f$ step horizon.

Intuitively, the problem can be mitigated by using errors of more steps, so that the model parameters are modified to better fit the ground-truth parameter at $\tau$ step before. However, human behavior is essentially time-varying. Too many steps may also create a gap between historic behavior and current behavior, so that the model adapted at an earlier time may be outdated and incapable to track the current behavior pattern. Thus there is indeed a trade-off between obtaining more information and maintaining behavior continuity when we increase observation steps $\tau$. The best step highly depends on the data distribution and model space, which is empirically analyzed in Sec.\ref{exp:adaptation}.

\begin{algorithm}[t]
	\caption{$\tau$ step online adaptation with MEKF$_{\lambda}$}
	\label{alg:adaptation}
		\textbf{Input:}
		Offline trained EDN network with parameter $\theta$\newline
		\textbf{Output:}
		A sequence of generated future behavior $\{\hat{\textbf{Y}}_{t}\}_{t=1}^T$
	\begin{algorithmic}[1]
		\For{$t=1, 2, ..., T$} 
		    \If {$t\geq\tau$}
		        \State stack recent $\tau$-step observations:
		       \State\hspace{2em}$\textbf{Y}_{t-\tau, t} = [y_{t-\tau+1}, ..., y_{t}]$
		        
		        \State stack recent $\tau$-step generated trajectory:
		         \State\hspace{2em}$\hat{\textbf{Y}}_{t-\tau, t} = [\hat{y}_{t-\tau+1}, ..., \hat{y}_{t}]$
		        \State adapt model parameter via MEKF$_{\lambda}$: 
		       \State\hspace{2em}$\hat{\theta_t}=f_{MEKF_{\lambda}}(\theta_{t-1}, \hat{\textbf{Y}}_{t-\tau, t}, \textbf{Y}_{t-\tau, t})$
		    \Else
		        \State initialization: $\hat{\theta}_t=\theta$
		    \EndIf
		    \State collect goal state $g_t$ from SGN and input features $\textbf{S}_t$.
		    \State generate future behavior:
		        \State\hspace{2em}$\hat{\textbf{Y}}_{t}=[\hat{y}_{t+1}, ...\hat{y}_{t+T_f}]=f_{EDN}(\textbf{S}_t, g_t, \hat{\theta}_t)$.
    	 \EndFor
	 \State\Return sequence of behaviors $\{\hat{\textbf{Y}}_{t}\}_{t=1}^T$
	\end{algorithmic}
\end{algorithm}

\subsection{Robust nonlinear adaptation algorithms}
\label{sec:MEKF}
There are many online adaptation approaches, such as stochastic gradient descent (SGD)\cite{bhasin2012robust}, recursive least square parameter adaptation algorithm (RLS-PAA)\cite{ljung1991result}. In this paper, we choose the modified extended Kalman filter with forgetting factors (MEKF$_{\lambda}$) \cite{abuduweili2021robust} as the adaptation algorithm. 

The MEKF$_{\lambda}$ regards the adaptation of a neural network as a parameter estimation process of a nonlinear system with noise:
\begin{equation}
    \textbf{Y}_{t} = f_{EDN}(\hat{\theta}_{t}, \textbf{S}_t) + \textbf{u}_t    
\end{equation}
\begin{equation}
    \hat{\theta}_{t} = \hat{\theta}_{t-1} + \omega_{t}
\end{equation}
where $\textbf{Y}_{t}$ is the observation of the ground-truth trajectory; $\hat{\textbf{Y}}_{t} = f_{EDN}(\hat{\theta}_{t}, \textbf{S}_t)$ is the generated behavior by the EDN policy $f_{EDN}$ with the input $\textbf{S}_t$  at time step $t$; $\hat{\theta}_t$ is the estimate of the model parameter of the EDN;
the measurement noise $u_{t}\sim \mathcal{N}(0, \textbf{R}_t)$ and the process noise $\omega_{t}\sim \mathcal{N}(0, \textbf{Q}_t)$ are assumed to be Gaussian with zero mean and white noise, namely $\textbf{R}_t$ and $\textbf{Q}_t$. Since the correlation among noises are unknown, it is reasonable to assume they are identical and independent of each other. For simplicity, we assume $\textbf{Q}_t=\sigma_q\textbf{I}$ and $\textbf{R}_t=\sigma_r\textbf{I}$ where $\sigma_q>0$ and $\sigma_r>0$. Applying MEKF$_{\lambda}$ on the above dynamic equations, we obtain the following equations to update the estimate of the model parameter:
\begin{equation}
    \hat{\theta}_{t} = \hat{\theta}_{t-1} + \textbf{K}_t \cdot (\textbf{Y}_t - \hat{\textbf{Y}_t})
\end{equation}
\begin{equation}
    \textbf{K}_t = \textbf{P}_{t-1}\cdot\textbf{H}_t^T\cdot(\textbf{H}_t\cdot\textbf{P}_{t-1}\cdot\textbf{H}_t^T+\textbf{R}_t)^{-1} 
\end{equation}
\begin{equation}
    \textbf{P}_t=\lambda^{-1}(\textbf{P}_t-\textbf{K}_t\cdot\textbf{H}_t\cdot\textbf{P}_{t-1}+\textbf{Q}_t)
\end{equation}
where $\textbf{K}_t$ is the Kalman gain. $\textbf{P}_t$ is a matrix representing the uncertainty in the estimates of the parameter $\theta$ of the model; $\lambda$ is the forgetting factor to discount old measurements; $\textbf{H}_t$ is the gradient matrix by linearizing the network:
\begin{equation}
    \textbf{H}_t = \frac{\partial f_{EDN}(\hat{\theta}_{t-1}, \textbf{S}_{t-1})}{\partial \hat{\theta}_{t-1}} = \frac{\partial\hat{\textbf{Y}}_{t-1}}{\partial \hat{\theta}_{t-1}}
\end{equation}
In implementation, we need to specify initial conditions $\theta_0$ and $\textbf{P}_0$. $\theta_0$ is initialized by the offline trained model parameter. For $\textbf{P}_0$, due to absence of prior knowledge on the initial model parameter uncertainty, we simply set it as an identity matrix $\textbf{P}_0=p_i\textbf{I}$ with $p_i>0$. Besides, EMKF$_{\lambda}$ enables us to adapt the parameter of different layers to find the best performance. The whole process of the online adaptation is summarized in Algorithm \ref{alg:adaptation} and illustrated in Figure~\ref{fig:adaptation_trajectory}.

%% file: 7experiment_detail.tex
\section{Detailed experiment}
\label{appendix:experiment}

\subsection{Semantic graph network evaluation}
\begin{table*}[t]
		\centering
		\caption{The statistical evaluation of the high-level intention identification policy. We compared our method with other six approaches to explore the effect of different representations and architecture.} 
		\label{tab:graph}
		\resizebox{\columnwidth}{!}{%
		\begin{tabular}{cc|ccc|ccc|c} 
			\toprule 
			& & \multicolumn{3}{c|}{Representation Ablation Study} & \multicolumn{3}{c|}{Architecture Ablation Study}& \\
			\midrule 
			Scenario & Measure & No-Temporal & GAT & Single-Agent & Two-Layer-Graph & Multi-Head & Seq-Graph & Ours\\
			\midrule 
			\multirow{2}*{Intersection} &  Acc (\%) & 87.44 ± 33.13 & \textbf{91.93 ± 27.05} & 88.8 ± 31.53 & 90.15 ± 29.79 & 90.00 ± 30.00 & 89.8 ± 30.24 & 90.50 ± 28.70\\
			~ & ADE (m) & 1.59 ± 1.67 & 1.18 ± 1.51 & 1.04 ± 0.90 & 0.98 ± 0.93 & 0.97 ± 0.75 & 1.33 ± 1.91 & \textbf{0.94 ± 0.73}\\
			\midrule    
			
			\multirow{2}*{\tabincell{c}{Roundabout\\(Transfer)}} &  Acc (\%) & \textbf{93.92 ± 23.88} & 91.21 ± 28.12 & 92.20 ± 26.75 & 90.54 ± 29.26 & 92.10 ± 26.96 & 91.60 ± 27.62 & 90.70 ± 29.08\\
			~ & ADE (m) & 3.62 ± 6.72 & 2.70 ± 5.16 & 1.88 ± 2.48 & 1.87 ± 2.51 & 2.79 ± 20.78 & 3.10 ± 3.10 & \textbf{1.70 ± 1.99}\\
			\bottomrule 
		\end{tabular} %
		}

\end{table*}

In the high-level intention-identification task, we compared the performance of our semantic graph network (SGN) with that of other six approaches. Three of them are exploring the effect of different representation of the input and output. And the rest of them are the variants of the proposed network, which explored the effect of frequently used network architectures and tricks.

\begin{enumerate}
    \item No-Temporal: This method does not take historic information into account, namely only consider the information of current time step $t$.
    \item GAT: This method uses absolute feature to calculate relationships among nodes, and no relative feature is defined. This method corresponds to the original graph attention network\cite{velivckovic2017graph}.
    \item Single-Agent: This method only considers the loss of ego vehicle's intention prediction, and does not consider the intention prediction for other vehicles.
    \item Two-Layer-Graph: This method has a two-layer graph to conduct information embedding, namely exploits the graph aggregations twice\cite{sanchez2018graph}.
    \item Multi-Head: This method employs the multi-head attention mechanism to stabilize learning\cite{velivckovic2017graph}. This method operates the relationship reasoning in Sec.\ref{sec:Relationship Reasoning Layer} multiple times in parallel independently, and concatenates all aggregated features as the final aggregated feature. In our case, we set the head number as 3.  
    \item Seq-Graph: This methods first conducts relationship reasoning for the graph at each time step and second feeds the sequence of aggregated graphs into RNN for temporal processing. As a comparison, our method first embeds each node's sequence of historic features with RNN and second conduct relationship reasoning using each node's hidden state from RNN at current time. 
\end{enumerate}

The models were trained and tested on the intersection scenario. The trained models were also directly tested on the roundabout scenario, which aimed at evaluating the zero-shot transferability. The performance of predicting which area to insert into was evaluated by calculating the multi-class classification accuracy. The performance of goal state generation was evaluated by the absolute-distance-error (ADE) between the generated goal state and ground-truth state.

According to the result shown in Table~\ref{tab:graph}, for the \textbf{insertion identification task}, all models achieved close accuracy of around 90\%, generally benefit from our representation of semantic graph. Another big view is that most models' transferability performance in the roundabout scenario surprisingly surpassed the performance in the intersection scenario which the models were originally trained. This is basically because intersection is a harder scenario than roundabout, as the vehicles need to interact with many vehicles from different directions simultaneously when they are entering the intersection, while vehicles in the roundabout only need to interact with the cars either from nearby branches or in close distance.

By a detailed analysis for insertion accuracy, though the model's performance were close, the GAT had the highest performance while our method followed as the second. The No-Temporal method had the lowest accuracy and largest variance in intersection scenario (87.44±22.13\%). This is because it is lacking in the temporal information, which could otherwise efficiently help to identify which DIA to insert into by introducing historic speed and acceleration. What is interesting is that the No-Temporal method contrarily achieved the highest insertion accuracy (93.92±23.88) in the roundabout scenario. One possible explanation is that the absence of temporal information constrained the model's capability and thus avoided over-specification, so the No-Temporal method has the best transferability performance. Such guess also helps to explain why the Two-Layer-Graph method had the lowest insertion accuracy in the roundabout scenario (90.54±29.26\%), as twice aggregations makes the model brittle to over-specification. 

The \textbf{goal state generation task}, on the one hand, is practically more important as it is directly delivered to low-level policy to guide the behavior generation process. On the other hand, it is much harder than the insertion identification task as it requires more delicate information extraction and inference. Consequently, we can see that the performance of different models varied a lot. Also, the models' performance significantly down-graded when they were directly transferred to the roundabout scenario, as the two scenarios have different geometries and different driving patterns such as speed and steering. 

Specifically, we have several observations: 1) our method achieved the lowest error in both intersection and roundabout scenarios; 2) the No-Temporal method was the worst in both intersection and roundabout scenario, due to the lack of temporal information; 3) the GAT method generated higher errors than our method especially in the roundabout scenario, which shows the necessity of the relative features as it might be helpful for better inference and distribution shift resistance. 4) Our method outperformed the Single-Agent method, which implies the advantages of data augmentation and encouraging interaction inference by taking all vehicles' generated goal state into the loss function. 5) The Two-Layer-Graph method 
was the closest one to our method, while it came with serious over-fitting according to our training log. 6) The Multi-Head method achieved the second best accuracy in intersection scenario but much worse performance in roundabout scenario, which could be possibly improved by careful tuning or searching for a proper head number. 7) The Seq-Graph method was the second worst in both the intersection and the roundabout scenario, which may imply that the complex encoding for past interactions could hardly help prediction but indeed makes the learning harder.


\begin{table}[!t]
		\centering
		\caption{The statistical evaluate of the low-level behavior generation policy. We conducted experiments incrementally to explore the performance under different coordinates, features, representations, and mechanisms.} 
		\label{tab:encoder_decoder}
		\begin{tabular}{cc|cc|cc} 
			\toprule 
            \multicolumn{2}{c|}{\multirow{2}*{\tabincell{c}{(a) Coordinate\\Study}}} &
			\multicolumn{2}{c|}{Intersection} & \multicolumn{2}{c}{Roundabout (Transfer)}\\
			
			\cmidrule(r){3-6} 
			~ & ~ & ADE & FDE & ADE & FDE\\
			\midrule 
			\multicolumn{2}{c|}{Cartesian} & 1.53 ± 1.22 & 2.90 ± 2.77 & 12.57 ± 5.68 & 19.77 ± 7.43 \\
			\multicolumn{2}{c|}{\textcolor{red}{Frenet}} & \textbf{0.91 ± 0.59}
            & \textbf{1.87 ± 1.48} & \textbf{2.96 ± 4.65} & \textbf{5.52 ± 7.20} \\

			\midrule 
			\bottomrule 
		\end{tabular} 
		\vspace{1em}

		\begin{tabular}{cc|cc|cc} 
			\toprule 
			\multicolumn{2}{c|}{\tabincell{c}{(b) Speed\\Ablation}} &
			\multicolumn{2}{c|}{Intersection} & \multicolumn{2}{c}{Roundabout (Transfer)}\\
			
			\midrule 
			\tabincell{c}{In} &\tabincell{c}{Out}
			&ADE & FDE & ADE & FDE\\
			\midrule 
			× & × & 0.91 ± 0.59 & 1.87 ± 1.48 & 2.96 ± 4.65 & 5.52 ± 7.20 \\
			\textcolor{red}{\checkmark} & \textcolor{red}{×} & 0.71 ± 0.54
            & 1.53 ± 1.27 & \textbf{2.33 ± 3.84} & \textbf{4.30 ± 5.44} \\
			× & \checkmark & 0.78 ± 0.53 & 1.74 ± 1.41 & 2.51 ± 4.16 & 4.97 ± 6.35 \\
			\checkmark & \checkmark & \textbf{0.70 ± 0.49} & \textbf{1.46 ± 1.26} & 2.40 ± 4.10 & 4.67 ± 6.03 \\
			
			\midrule 
			\bottomrule 
		\end{tabular} 
		\vspace{1em}
		
		\begin{tabular}{cc|cc|cc} 
			\toprule 
			\multicolumn{2}{c|}{\tabincell{c}{(c) Yaw\\Ablation}} &
			\multicolumn{2}{c|}{Intersection} & \multicolumn{2}{c}{Roundabout (Transfer)}\\
			
			\midrule 
			\tabincell{c}{In} &\tabincell{c}{Out}
			&ADE & FDE & ADE & FDE\\
			\midrule 
			× & × & 0.71 ± 0.54 & 1.53 ± 1.27 & 2.33 ± 3.84 & 4.30 ± 5.44 \\
			\textcolor{red}{\checkmark} & \textcolor{red}{×} & \textbf{0.67 ± 0.46} & \textbf{1.45 ± 1.19} & \textbf{2.23 ± 3.95} & \textbf{4.14 ± 5.19} \\
			× & \checkmark & 0.73 ± 0.50 & 1.59 ± 1.36 & 2.34 ± 3.96 & 4.49 ± 6.46 \\
			\checkmark & \checkmark & 0.67 ± 0.48 & 1.46 ± 1.24 & 2.51 ± 4.60 & 4.77 ± 6.43 \\
			
			\midrule 
			\bottomrule 
		\end{tabular} 
		\vspace{1em}

			
		
		\begin{tabular}{cc|cc|cc} 
			\toprule 
			\multicolumn{2}{c|}{\tabincell{c}{(d) Repre\\Ablation}} &
			\multicolumn{2}{c|}{Intersection} & \multicolumn{2}{c}{Roundabout (Transfer)}\\
			
			\midrule 
			\tabincell{c}{Inc} &\tabincell{c}{Ali}
			&ADE & FDE & ADE & FDE\\
			\midrule 
			× & × & 0.67 ± 0.46 & 1.45 ± 1.19 & 2.23 ± 3.95 & 4.14 ± 5.19 \\
			× & \checkmark & 0.48 ± 0.44 & 1.32 ± 1.32 & 1.26 ± 0.95 & 3.04 ± 2.44 \\
			\checkmark & × & 0.43 ± 0.35 & 1.36 ± 1.19 & 1.07 ± 1.10 & 2.66 ± 2.31 \\
			\textcolor{red}{\checkmark} & \textcolor{red}{\checkmark} & \textbf{0.41 ± 0.33} & \textbf{1.29 ± 1.14} & \textbf{0.96 ± 0.80} & \textbf{2.53 ± 2.13} \\
			
			\midrule 
			\bottomrule 
		\end{tabular} 
		\vspace{1em}
		
		\begin{tabular}{cc|cc|cc} 
			\toprule 
			\multicolumn{2}{c|}{\tabincell{c}{(e) Mech\\Ablation}} &
			\multicolumn{2}{c|}{Intersection} & \multicolumn{2}{c}{Roundabout (Transfer)}\\
			
			\midrule 
			\tabincell{c}{TF} &\tabincell{c}{Att}
			&ADE & FDE & ADE & FDE\\
			\midrule 
			\textcolor{red}{×} & \textcolor{red}{×} & \textbf{0.41 ± 0.33} & \textbf{1.29 ± 1.14} & \textbf{0.96 ± 0.80} & \textbf{2.53 ± 2.13} \\
			\checkmark & × & 0.41 ± 0.34 & 1.32 ± 1.16 & 0.97 ± 0.83 & 2.54 ± 2.14 \\
			× & \checkmark & 0.43 ± 0.34 & 1.32 ± 1.13 & 1.10 ± 1.32 & 2.57 ± 2.50 \\

			\midrule 
			\bottomrule 
		\end{tabular} 
		\vspace{1em}
		
\end{table}

\subsection{Encoder decoder network evaluation}
There are many existing works exploiting the encoder decoder architecture for the driving behavior generation\cite{park2018sequence}\cite{tang2019multiple}\cite{zyner2019naturalistic}, but several questions still remain unclear, such as what coodinate should be employed, what features should be considered, what representation performs better, and whether commonly-used mechanisms in encoder decoder architecture can improve the performance in the driving task. To answer these questions, we experimented with the EDN itself without intention signal. Note that, starting from a naive encoder decoder which simply takes in position features and predicts positions, we conducted the experiments incrementally. Later experiments inherited the structure verified as the best earlier. We set two metrics, absolute distance error (ADE) and final distance error (FDE).

\subsubsection{Coordinate study}
we first investigated which coordinate should we employ between Frenet and Cartesian coordinate. As shown in Table~\ref{tab:encoder_decoder}(a), we can see the EDN with Frenet coordinate performed 40\% (ADE) and 35\% (FDE) better than EDN with Cartesian coordinate in the intersection scenario. In the zero-transferred roundabout scenario, though the performance of both two methods downgraded, the performance of method with Cartesian coordinate decayed more significantly, with ADE higher by 324\%  and FDE higher by 258\% compared to the method with Frenet coordinate.

This is because the Frenet coordinate implicitly incorporate the map information into the model. Compared to running in any directions in the Cartesian coordinate, the vehicles would follow the direction of references paths in the Frenet coordinate, which constrains its behavior in a reasonable range.

\subsubsection{Feature study}
Second, we explored the effect of features, specifically, the speed feature and yaw feature. For each feature, we consider two circumstances. The first is to incorporate the feature into the input of the encoder to provide more information. The second is to set the feature as additional desired outputs of the decoder, which could possibly help to stabilize the learning for position prediction. Thus for each feature, we explored 4 settings in terms of whether or not adding the feature into the input or output.

As in Table~\ref{tab:encoder_decoder}(b), incorporating speed feature in either the input or output could both effectively improve performance in the two scenarios. When incorporating it into the input and output simultaneously, the performance was slightly improved in the intersection scenario and slightly degraded in the roundabout scenario compared to only considering it in the input. Judging from the average performance in two scenarios, we choose to only take the speed into the input of encoder.

For the yaw feature, as shown in Table~\ref{tab:encoder_decoder}(c), the approach taking it into input could slightly benefit the performance, while incorporating it into the output made the performance worse. One possible reason for such performance decay is that the yaw information has been already implicitly covered in the speed information. Not providing additional information, adding the yaw information into the output of decoder indeed made the learning harder. We thus decided to incorporate the yaw feature into the input of the encoder.

\subsubsection{Representation study}
There are two commonly-used techniques to shape the data distribution. The first technique is called incremental prediction\cite{li2019grip++}, which predicts position difference compared to the position of last step, rather than directly predict the absolute position. The second technique is called position alignment, which align the positions of each steps to the vehicle's current position\cite{park2018sequence}.

According to Table~\ref{tab:encoder_decoder}(d), both two techniques could significantly improve the prediction accuracy, while applying both of them worked the best, improving the ADE by 38\% in the intersection and by 56\% in the roundabout. 

\subsubsection{Mechanism study}
There are two frequently used mechanisms in the encoder decoder architecture: teacher forcing\cite{williams1989learning} and attention mechanism\cite{bahdanau2014neural}. Teacher forcing aims at facilitating the learning of complex tasks while the attention is designed to attend to different historic input. 

From the results in Table~\ref{tab:encoder_decoder}(c), we can see neither of the two mechanisms could benefit our model. Considering that the encoder decoder is used as a dynamics approximator, which is a relatively simple task, the teacher forcing achieved similar performance as the EDN itself can already learn well enough. The attention mechanism indeed made the performance worse as the vehicle dynamics is most related to recent state and previous states may not be necessarily relevant. 

\subsection{Intention signal integration evaluation}
\label{sec:experiment_intention_signal_intro}
As mentioned in Sec. \ref{sec:intention_signal_intro_method}, we have two intention signals, namely the goal state and the decoding step, which can be integrated into the EDN in several ways, such as appending it into the input or output of the decoder (note as Input and Output), embedding it into the hidden state at the first step (note as Hidden). For the goal state, we can additionally choose to introduce it by transforming the origin state of the vehicle into the state relative to the goal state (note as Transform). 

First, we introduced the ground-truth goal state into the EDN to verify whether the ground-truth intention could benefit prediction. According to Table~\ref{tab:intention}(a), the Transform method had the best performance and reduced the ADE by 75\% and 56\% in the two scenarios, which represents the most benefit we can get with ground-truth intention but is also impossible as there exists unevitable errors in the predicted goal state. Adding goal state into the hidden state also outperformed appending it into the input, while adding it at the output of GRU performed worst.

When integrating the predicted goal state into EDN, the error in the predicted goal state would perturb the performance. As in Table~\ref{tab:intention}(b), while the performance of these goal state integration methods were close, appending the goal state into the input feature list performed the best, reducing the ADE by 26\% and 10\% in the two scenarios.

After introducing goal state into the input feature, we further investigate how to introduce the time signal as in Table~\ref{tab:intention}(c). Similarly, appending it into the input performed the best, which reduced the error especially in roundabout scenario by 5\% compared to only considering the goal state signal.

In Figure~\ref{fig:Error by Step}, we illustrated the effect of introducing intention signal, by calculating the prediction error of each step in the future 30 steps. Obviously, as the prediction horizon extended, the prediction became more difficult and the error grew exponentially. When introducing the intention signal, the error growth was effectively suppressed, especially in the long horizon.

\begin{table*}[!t]
		
		\centering
		\caption{The statistical evaluation of different methods to introduce intention signals into the low-level trajectory prediction policy.} 
		\label{tab:intention}
		\begin{tabular}{cc|c|cccc} 
			\toprule 
			\multicolumn{7}{c}{(a) Ground-truth Goal State Introduction}\\
			\midrule 
			Scenario & Measure & No-Label & Transform & Input & Output & Hidden\\
			\midrule 
			\multirow{2}*{Intersection} &  ADE (m) & 0.41 ± 0.11 & \textbf{0.10 ± 0.10} & 0.15 ± 0.13 & 0.17 ± 0.16 & 0.13 ± 0.14\\
			~ & FDE (m) & 1.29 ± 1.30 & \textbf{0.15 ± 0.25} & 0.29 ± 0.40 & 0.38 ± 0.41 & 0.26 ± 0.39\\
			\midrule    
			
			\multirow{2}*{\tabincell{c}{Roundabout\\(Transfer)}} &  ADE (m) & 0.96 ± 0.64 & \textbf{0.42 ± 0.37} & 0.51 ± 0.46 & 0.60 ± 0.54 & 0.48 ± 0.41\\
			~ & FDE (m) & 2.53 ± 4.54 & \textbf{0.72 ± 0.66} & 0.94 ± 0.73 & 1.03 ± 0.74 & 0.84 ± 0.67\\
			\bottomrule 
		\end{tabular} 
		\vspace{1em}
		
		\begin{tabular}{cc|c|cccc} 
			\toprule 
			\multicolumn{7}{c}{(a) Generated Goal State Introduction}\\
			\midrule 
			Scenario & Measure & No-Label & Transform & Input & Output & Hidden\\
			\midrule 
			\multirow{2}*{Intersection} &  ADE (m) & 0.41 ± 0.11 & 0.31 ± 0.25 & \textbf{0.30 ± 0.25} & 0.32 ± 0.25 & 0.31 ± 0.25\\
			~ & FDE (m) & 1.29 ± 1.30 & 0.92 ± 0.85 & 0.89 ± 0.83 & 0.89 ± 0.83 & \textbf{0.89 ± 0.82}\\
			\midrule    
			
			\multirow{2}*{\tabincell{c}{Roundabout\\(Transfer)}} &  ADE (m) & 0.96 ± 0.64 & 0.89 ± 0.56 & \textbf{0.86 ± 0.61} & 0.92 ± 0.75 & 0.87 ± 0.54\\
			~ & FDE (m) & 2.53 ± 4.54 & 2.14 ± 1.44 & \textbf{2.12 ± 1.51} & 2.19 ± 1.69 & 2.22 ± 1.46\\
			\bottomrule 
		\end{tabular} 
		\vspace{1em}

		\begin{tabular}{cc|c|cccc} 
			\toprule 
			\multicolumn{7}{c}{(a) Time Signal Introduction}\\
			\midrule 
			Scenario & Measure & No-Label & Transform & Input & Output & Hidden\\
			\midrule 
			\multirow{2}*{Intersection} &  ADE (m) & 0.41 ± 0.11 & 0.30 ± 0.25 & \textbf{0.30 ± 0.25} & 0.30 ± 0.25 & 0.30 ± 0.24\\
			~ & FDE (m) & 1.29 ± 1.30 & 0.89 ± 0.83 & 0.88 ± 0.83 & 0.89 ± 0.83 & \textbf{0.88 ± 0.81}\\
			\midrule    
			
			\multirow{2}*{\tabincell{c}{Roundabout\\(Transfer)}} &  ADE (m) & 0.96 ± 0.64 & 0.86 ± 0.61 & \textbf{0.82 ± 0.53} & 0.87 ± 0.59 & 0.84 ± 0.53\\
			~ & FDE (m) & 2.53 ± 4.54 & 2.12 ± 1.51 & \textbf{2.06 ± 1.43} & 2.15 ± 1.51 & 2.11 ± 
			0.23\\
			\bottomrule 
		\end{tabular} 
		
		\vspace{1em}
\end{table*}

\subsection{Online adaptation evaluation}
\label{exp:adaptation}
\begin{figure}[t]
    \centering
    \includegraphics[width=0.5\textwidth]{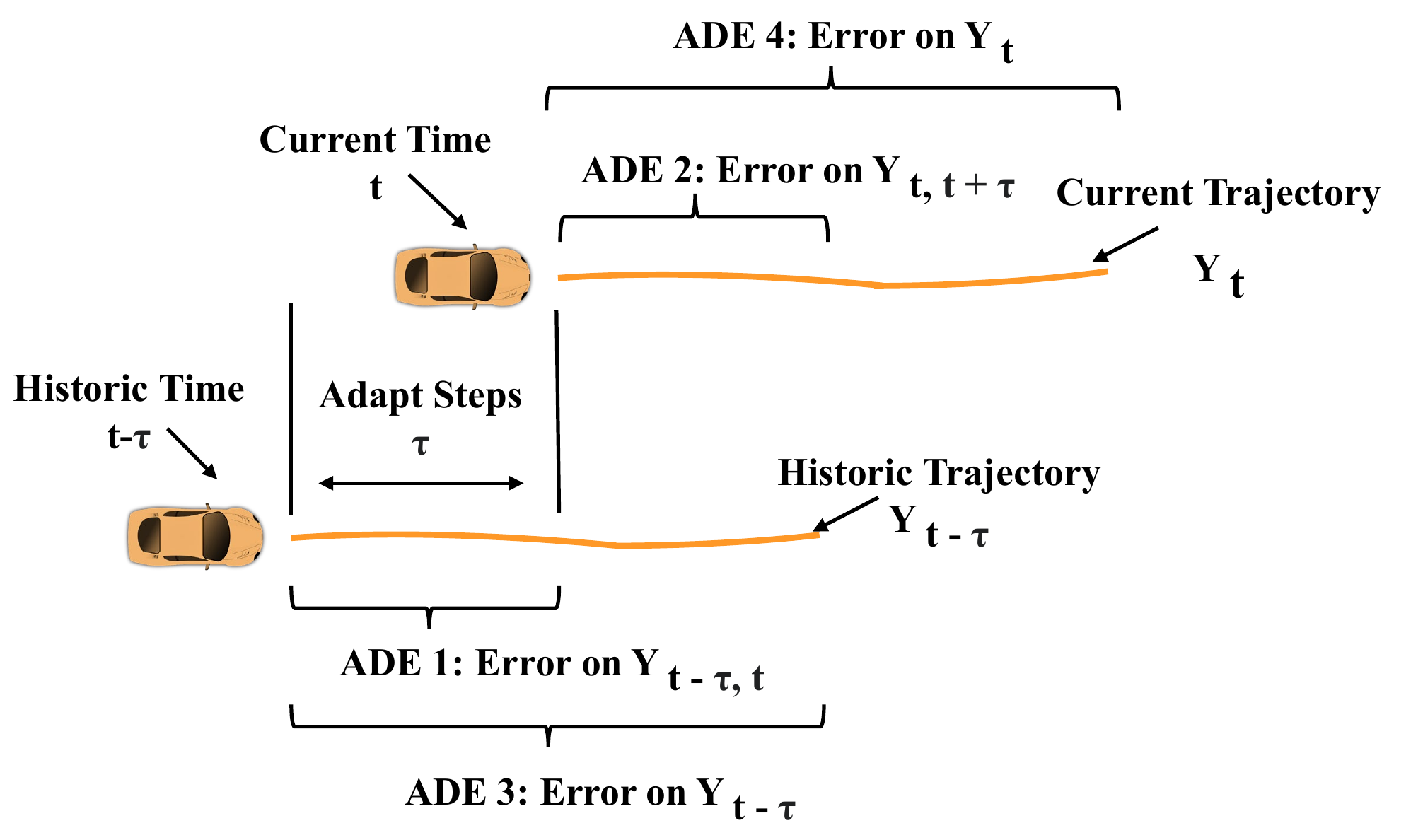}
    \caption{An illustration for online adaptation and 4 metrics for performance analysis. At time step $t$, the model parameters can be adapted by minimizing prediction error of the trajectory in past $\tau$ steps $Y_{t-\tau, t}$. 4 metrics are designed to analyze the adaptation performance. ADE 1 can verify how the adaptation works on the source trajectory $Y_{t-\tau, t}$. ADE 2 can verify whether adaptation can benefit short-term prediction in the presence of behavior gap between earlier time and current time. ADE 3 can verify whether we have obtained enough information. ADE 4 verifies whether adaptation can benefit the long-term prediction.}
    \label{fig:adaptation_trajectory}
\end{figure}

As in Figure \ref{fig:adaptation_trajectory}, we designed several metrics to systematically analyze the performance of online adaptation:
\begin{enumerate}
    \item ADE 1: This metric evaluates the prediction error of the adapted steps on the historic trajectory $Y_{t-\tau, t}$. Because these steps are the observation source used to conduct online adaptation, the metric can verify whether the algorithm is working or not.
    \item ADE 2: This metric evaluates the prediction error of the adaptation steps on the current trajectory, which aims at verifying how the time lag is influencing the prediction. Also, this method can be used to verify whether adaptation could improve short-term behavior prediction.
    \item ADE 3: This metric evaluate the prediction error of the whole historic trajectory, which implies whether we have get enough information on the behavior pattern.
    \item ADE 4: This metric evaluates the performance of the whole current trajectory, which shows whether the adaptation based on historic information could help current long-term behavior generation.
\end{enumerate}

The adaptation step $\tau$ is an important parameter in the online adaptation algorithm. On the one hand, we can obtain more information by increasing $\tau$. On the other hand, the behavior gap between current time and historic time also increased. As a result, there is a performance trade-off when we increase the adaptation step $\tau$. To empirically answer the question that how many steps are the best, We run the online adaptation on both the intersection and roundabout scenarios, and collected statistical results of the ADE between ground-truth and predicted behavior. Note that here we adapted different the parameter of different layers and chose the best adaptation performance. 

\begin{figure*}[t!]
    \centering
    \includegraphics[width=1\textwidth]{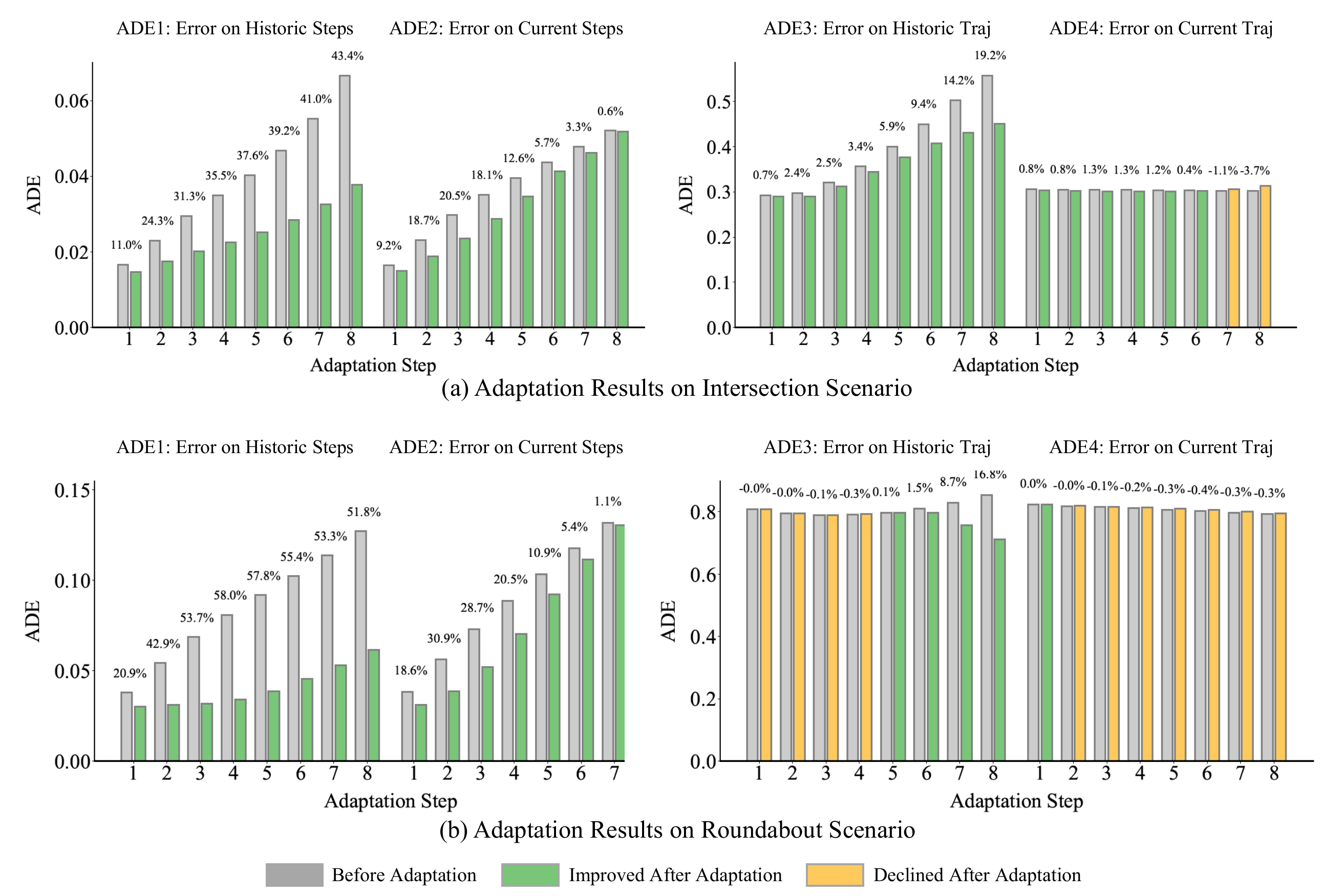}
    \caption{Adaptation Analysis. According to ADE 1 and ADE 3 in (a)(b), as adaptation step $\tau$ increased, the online adaptation could get more information and improve prediction by a higher percentage. In the ADE 2, as adaptation step $\tau$ increases, the improved percentage first grew higher but then declined, due to the trade-off between more information and behavior gap. When $\tau$ was 3, the short-term prediction was improved by 20\% and 28\% on the two scenarios respectively. In the ADE 4, we can see online adaptation can barely benefit long-term prediction.}
    \label{fig:adaptation_analysis}
\end{figure*}

Figure~\ref{fig:adaptation_analysis}(a) shows the online adaptation results in the intersection scenario. According to the ADE 1 and ADE 2 in the firt two images, as the adaptation step $\tau$ increased, the prediction error itself of the first $\tau$ step increased for both historic trajectory (ADE 1) and current trajectory (ADE 2). Also, for ADE 1, the percentage of error reduction increased along with the increasing adaptation step $\tau$, because more information was gained. For the ADE 2, as the adaptation steps increased, the percentage of error reduction first increased and reached a peak of 20.5\% at 3 adaptation step. After that, the percentage of error reduction decreased as the behavior gap had come into effect due to longer time lag. The last two images show the error in the whole historic trajectory (ADE 3) and current trajectory (ADE 4). For the ADE 3 in the third image, longer $\tau$ led to higher percentage of error reduction because of more information gained. However, for the ADE 4 in the fourth image, due to the insufficient information and behavior gap, the improvement in the long-term prediction was limited. 

Similar results can be found in the roundabout scenario in Figure~\ref{fig:adaptation_analysis}(b). But in the ADE 2, more improvement (28\%) was achieved in the short-term prediction, due to the fact that the model was not trained on the roundabout scenario and there was more space for adaptation.

With these analysis, a conclusion is that though the adaptation can not help with the long-term behavior prediction in next 3 seconds, the short-term behavior prediction in the next 0.3 seconds is effectively improved by 20.5\% and 28.7\% in the two scenarios. Such improvement in short-term prediction is valuable as it can effectively enhance safety in close-distance interactions.

%% file: 11appendix.tex
\section{Network architecture}


\begin{table}[h!]
	\centering
	\caption{Architecture detail for SGN and EDN.}
	\label{tab:GNN table}
	\begin{tabular}{c|c} 
		\toprule 
		\multicolumn{2}{c}{SGN Architecture Detail}\\
		\midrule 
		$f^1_{rec}$ & GRU cell\\
		$f^2_{rec}$ & GRU cell\\
		$f^1_{enc}$ & Dense layer with tanh activation function\\
		$f^2_{enc}$ & Dense layer with tanh activation function\\
		$f_{att}$ & Dense layer with leaky relu activation function\\
		$f^3_{enc}$ & Dense layer without activation function\\
		$f^4_{enc}$ & Dense layer with tanh activation function\\
		\bottomrule 
	\end{tabular} 
	\vspace{1em}
	
	\begin{tabular}{c|c} 
		\toprule 
		\multicolumn{2}{c}{Encoder Decoder Architecture Detail}\\
		\midrule 
		Encoder & GRU cell\\
		\midrule 
		Decoder & \tabincell{c}{GRU cell stacked with 3 dense layers, \\ each layer with tanh activation function and dropout}\\
		\bottomrule 
	\end{tabular} 
	
\end{table}